\documentclass[10pt,twocolumn,letterpaper]{article}
\usepackage[pagenumbers]{cvpr} 
\definecolor{cvprblue}{rgb}{0.21,0.49,0.74}
\usepackage[pagebackref,breaklinks,colorlinks,allcolors=cvprblue]{hyperref}
\usepackage{xcolor}         
\usepackage{graphicx}
\usepackage{bm}
\usepackage{algorithmic}
\usepackage{pgffor}
\usepackage{amsmath}
\usepackage{multirow}
\usepackage[misc]{ifsym}
\usepackage[accsupp]{axessibility}  
\usepackage[ruled,vlined,linesnumbered]{algorithm2e}

\usepackage{array}
\newcommand{\PreserveBackslash}[1]{\let\temp=\\#1\let\\=\temp}
\newcolumntype{C}[1]{>{\PreserveBackslash\centering}p{#1}}
\newcolumntype{R}[1]{>{\PreserveBackslash\raggedleft}p{#1}}
\newcolumntype{L}[1]{>{\PreserveBackslash\raggedright}p{#1}}

\title{FOLDER: Accelerating Multi-modal Large Language Models with Enhanced Performance}

\author{
Haicheng Wang$^{1,2*}$, 
Zhemeng Yu$^{1,2*}$, 
Gabriele Spadaro$^{2,3}$, 
Chen Ju$^{4}$,  \\
Victor Quétu$^{2}$,  
Shuai Xiao$^{4}$, 
Enzo Tartaglione$^{2}\textsuperscript{\Letter}$ 
\\
$^1$ SJTU Paris Elite Insitute of Technology, Shanghai Jiao Tong University, China \\
$^2$  LTCI, Télécom Paris, Institut Polytechnique de Paris, France \\
$^3$ University of Turin, Italy \ \
$^4$ Alibaba Group, China
\\
{\tt\small \{anakin\_skywalker,fish\_meng\}@sjtu.edu.cn, gabriele.spadaro@unito.it,} \\
{\tt\small
cju.void@gmail.com,
\{victor.quétu,enzo.tartaglione\}@telecom-paris.fr
} \\
\url{https://github.com/anakin-skywalker-Joseph/Folder}
}

\begin{document}
\maketitle
\begingroup
\renewcommand\thefootnote{\relax}\footnotetext{*: Equation contribution. \quad \textsuperscript{\Letter}: Corresponding author.}
\endgroup
\begin{abstract}
Recently, Multi-modal Large Language Models (MLLMs) have shown remarkable effectiveness for multi-modal tasks due to their abilities to generate and understand cross-modal data. However, processing long sequences of visual tokens extracted from visual backbones poses a challenge for deployment in real-time applications. To address this issue, we introduce \textbf{FOLDER}, a simple yet effective plug-and-play module designed to reduce the length of the visual token sequence, mitigating both computational and memory demands during training and inference. Through a comprehensive analysis of the token reduction process, we analyze the information loss introduced by different reduction strategies and develop FOLDER to preserve key information while removing visual redundancy. We showcase the effectiveness of FOLDER by integrating it into the visual backbone of several MLLMs, significantly accelerating the inference phase. Furthermore, we evaluate its utility as a training accelerator or even performance booster for MLLMs. In both contexts, FOLDER achieves comparable or even better performance than the original models, while dramatically reducing complexity by removing up to 70\% of visual tokens. 
\end{abstract}    
\vspace{-0.3cm}
\section{Introduction}
\label{sec:intro}

Multi-modal Large Language Models (MLLMs) have become a powerful framework for multimodal tasks, playing a key role in applications such as image captioning~\cite{cheng2023beyond} and visual question answering~\cite{khan2023q}. By learning a joint representation of visual and textual information, state-of-the-art MLLMs like GPT-4V~\cite{achiam2023gpt} and Claude~3~\cite{claude3} have demonstrated great capabilities in understanding and even generating multimodal content. However, these models face concrete challenges in terms of computational efficiency, particularly when processing visual inputs.

\begin{figure}[t]
  \centering
  \includegraphics[width=\columnwidth]{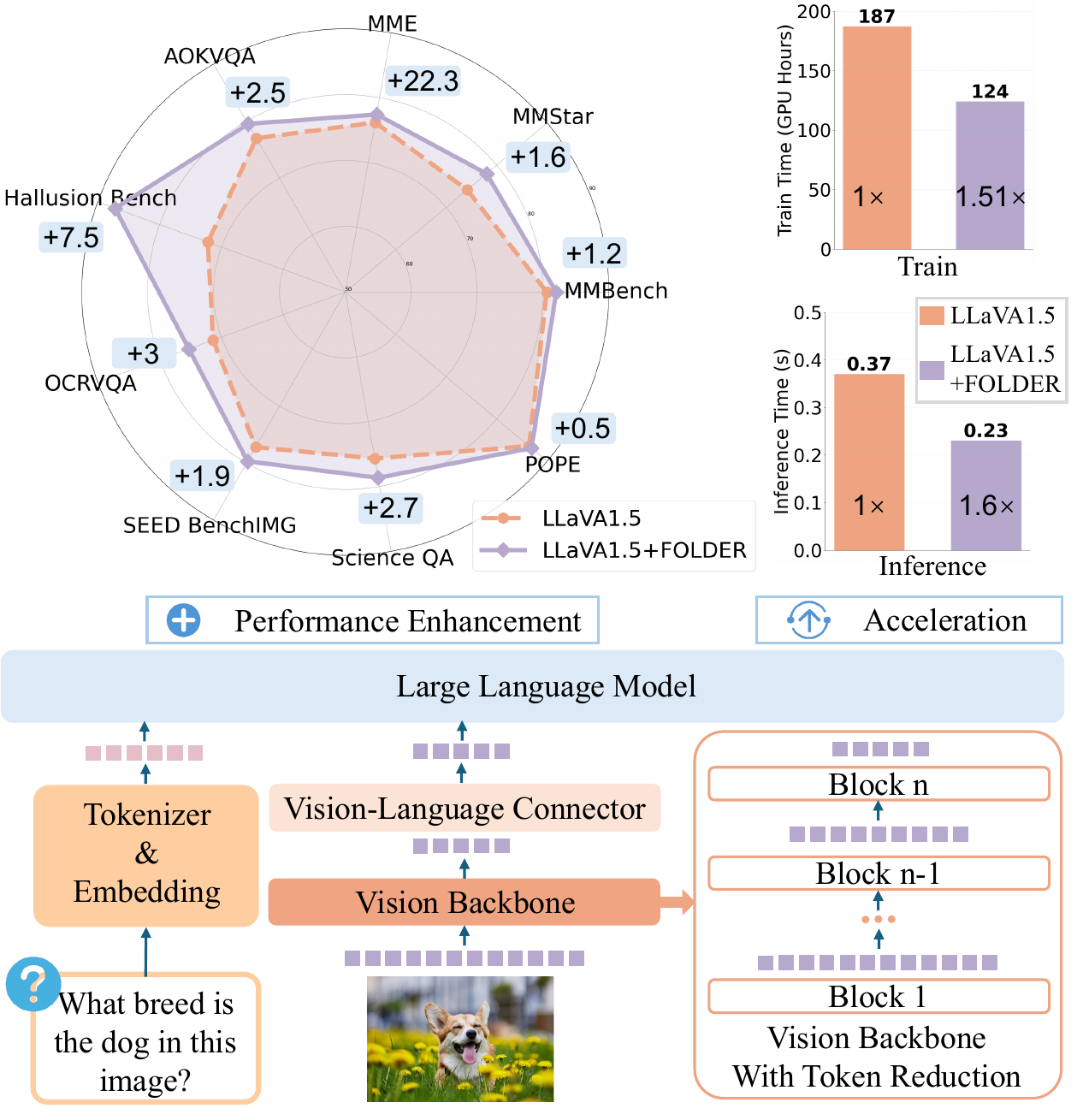}
  \caption{\textbf{FOLDER as Accelerator \& Booster.} As a plug-and-play module, FOLDER can be used in both training and inference, with considerable acceleration and even performance boost.}
  \label{fig:teaser}
  \vspace{-0.2cm}
\end{figure}

Modern MLLMs typically generate long sequences of image tokens through their visual backbones. In some application scenarios, this problem becomes even more evident: recent advancements in high-resolution-supporting MLLMs~\cite{wang2024qwen2,chen2024internvl} might involve over 2000 visual tokens. Similarly, multi-visual-expert architectures~\cite{shi2024eagle, kar2024brave, tong2024eyes} leverage multiple visual encoders, encountering analogous challenges. These issues are even more pronounced in video understanding models~\cite{li2023videochat, lin2023video, jin2024chat} due to the temporal dimension of video content. This rapid increase in token sequence length represents a challenge for real-time application deployment due to the \emph{quadratic computational complexity} of the attention mechanism. 

Several studies have been conducted to address this issue. Existing approaches can be roughly divided into two categories. The first category focuses on \emph{training phase} solutions, such as Q-Former~\cite{li2023blip}, resampler~\cite{chen2024internvl} and pooling-based techniques~\cite{chen2023minigpt}. However, these methods often suffer from performance degradation and exhibit limited scalability, due to their custom design for specific architectures. The second category aims to design plug-and-play token reduction modules for the \emph{inference phase}~\cite{ju2025turbo, chen2024image}. While these methods identify the inherent redundancy in visual tokens, their reduction strategies are often arbitrary, \eg employing a uniform reduction for each layer/block, resulting in sub-optimal outcomes. Specifically, these methods fail to consider the information loss during the token reduction process, thus leading to substantial performance drops.

To find a reduction strategy that can preserve the performance, we propose a preliminary investigation to answer a fundamental question: \emph{Where does information loss originate?} To model it, we identify three key factors that impact information loss during token removal: \textbf{Reduction Impact}, \textbf{Propagation Effect}, and \textbf{Aggregation Method}. Through a comprehensive empirical study of these factors, we devise a strategy that enhances the preservation of information while significantly reducing redundancy. FOLDER is consequently designed to implement this strategy with minimal computational overhead. 
Our plug-and-play solution can be seamlessly integrated into the visual backbones of various MLLMs, effectively reducing the number of visual tokens while incurring negligible information loss (Fig.~\ref{fig:teaser}). We first incorporate FOLDER during MLLMs inference, reducing over \emph{60\%} of visual tokens and achieving comparable, or even gains on performance across multiple tasks. Furthermore, when integrated into MLLMs pre-training, our module can be used as a training accelerator, as well as an effective regularization term, leading to remarkable performance improvements across all benchmarks, with a reduction ratio up to \emph{70\%}. 
Our method is on-the-field proven to be a flexible, nearly lossless, and highly efficient token reduction strategy, offering a dual-purpose solution to the visual token reduction challenge.
In summary, our contributions can be summarized as:
\begin{itemize}
    \item We conduct an in-depth analysis for the sources of information decay during token reduction, identifying and quantifying the key factors involved (Sec.~\ref{sec:reduction}, Sec.~\ref{sec:propagation} and Sec.~\ref{sec:aggregation}). These insights offer a clear understanding of how token reduction impacts the information flow. 
    \item Leveraging the above analysis, we develop a simple yet effective plug-and-play visual token reduction module for MLLMs (Sec.~\ref{sec:folder}). This novel token reduction strategy aggressively reduces the number of tokens only in the last blocks of the visual encoder. On various models and benchmarks, FOLDER speeds up the inference 1.7-2.4$\times$ with negligible loss or even improvement (Sec.~\ref{sec:inference_folder}).
    \item We demonstrate that our method can also be useful in MLLM training (Sec.~\ref{sec:train_folder}), improving both the performance and the speed.
\end{itemize}

\section{Related Works}
\label{sec:related work}

\vspace{0.1cm}
\noindent \textbf{Multi-modal Large Language Models} have gained considerable attention due to the powerful ability to understand multi-modality effectively~\cite{lxmert,visualbert,flava,achiam2023gpt,team2023gemini,llava,wang2024qwen2,zhang2023Video-LLaMA,li2023blip,chen2023minigpt,ataallah2024MiniGPT4-Video,lin2023video}, and can thus benefit lots of downstream tasks: image domain~\cite{wang2024advancing,ma2023diffusionseg,chen2024wear,ma2023open,yang2024multi,cheng2023image,ma2024open,chen2023enhancing,cheng2023mixer,yang2023multi,ma2023attrseg}, video domain~\cite{ju2022prompting,zhao2020bottom,ju2021divide,cheng2024denoiser,ju2021adaptive,ju2023multi,ju2023distilling,ju2020point,ju2023constraint}, as well as audio domain~\cite{liu2022exploiting,liu2024annotation,liu2024audio,liu2023audio}. 
A popular MLLM architecture for vision is composed of i.) Visual Backbone, ii.) Vision-Language Connector and iii.) Pre-trained Large Language Model. The integration of visual information leads to rapid growth in computational cost due to the large number of visual tokens. For instance, LLaVA1.5~\cite{liu2024improved} employs 576 tokens for a $336 \times 336$ image and up to several thousands of tokens for high-resolution images. This issue becomes even more obvious for video understanding models~\cite{li2023videochat,lin2023video,jin2024chat}. For instance, in VideoLLaVA~\cite{lin2023video}, even when performing inference with a limited 8 frames at a resolution of $224\times 224$, the sequence length already surpasses 2000 tokens.
Recently, some works~\cite{shi2024eagle,kar2024brave,tong2024eyes} demonstrate the importance of using multiple complementary vision towers to enhance MLLMs' visual ability. While effective, this approach even aggravates the sequence length issue.

\vspace{0.1cm}
\noindent \textbf{MLLMs Acceleration.} 
Many approaches focus on system-level optimizations for acceleration, such as FlashAttention~\cite{dao2022flashattention}, vLLM~\cite{dao2023flashattention2}, and RingAttention~\cite{liu2023ringAttention}. Techniques on knowledge distillation~\cite{wang2022efficientvlm,fang2021compressing}, quantization~\cite{xiao2023smoothquant,frantar2022gptq} and model pruning~\cite{wang2022efficientvlm,rao2021dynamicvit,shi2023upop} were also introduced to reduce model size and computational cost. However, these approaches fail to address the challenge of redundant data, which leads to unnecessarily long sequences.

\vspace{0.1cm}
\noindent \textbf{Token Reduction in MLLMs.} Some preliminary studies are conducted on token reduction for Vision Transformers~\cite{liang2022patchesneedexpeditingvision,kong2022spvitenablingfastervision,bolya2022token}. In the context of MLLMs, several token reduction methods, including Q-Former~\cite{li2023blip}, resampler~\cite{chen2024internvl} and pooling~\cite{chen2023minigpt} are proposed to reduce visual tokens during the training process. However, these approaches often suffer from performance degradation and lack of scalability.

Meanwhile, some studies try to handle the token reduction problem during inference. In particular, ToMe~\cite{bolya2022token} proposes a training-free token merging strategy for uni-modal image classification. Based on that, Turbo~\cite{ju2025turbo,ju2023turbo} proposes an improved merging strategy considering both mutual redundancy and semantic value, serving as a plug-in module in MLLMs. Recently, FastV~\cite{chen2024image} recognizes the visual attention sparsity in LLM, thus pruning visual tokens inside the LLM based on the task-orientated attention importance. Though effective, these methods have clear drawbacks. For ToMe/Turbo, the uniform progressive merging across all layers is arbitrary and inflexible, while FastV mainly suffers from two major problems: i.) FastV requires the attention map for every output token, so the kv-cache needs to keep the visual tokens throughout the whole dialogue, making it unsuitable for long-term visual QA. ii.) Since FastV is inserted in LLM, it is difficult to adapt for complex LLM modules~\cite{dao2022flashattention}. To address these issues, we seek to find an effective, information-guided, and user-friendly token reduction method that can be applied without constraints. Note that there are some concurrent works~\cite{shang2024llava,zhang2024sparsevlm,huang2024dynamic,yang2024visionzip} on MLLM token reduction.

\section{Method} \label{sec:method} 
In this section, we ground our FOLDER by first answering three key questions for token reduction: i.) how many tokens to reduce (Sec.~\ref{sec:reduction}); ii.) in which block (Sec.~\ref{sec:propagation}); iii.) through which aggregation method (Sec.~\ref{sec:aggregation}). Then, we present our method in Sec.~\ref{sec:folder}.

\vspace{0.1cm}
\noindent \textbf{Empirical Setup.} To answer these three crucial questions, we define a common empirical setup. Specifically, we use a pre-trained 12-block ViT-B~\cite{dosovitskiy2020image} model and conduct the empirical experiments on ImageNet-1k~\cite{deng2009imagenet}. Results on more diverse setups are provided in supplementary materials.

\begin{figure}[t]
\centering
\includegraphics[width=\columnwidth]{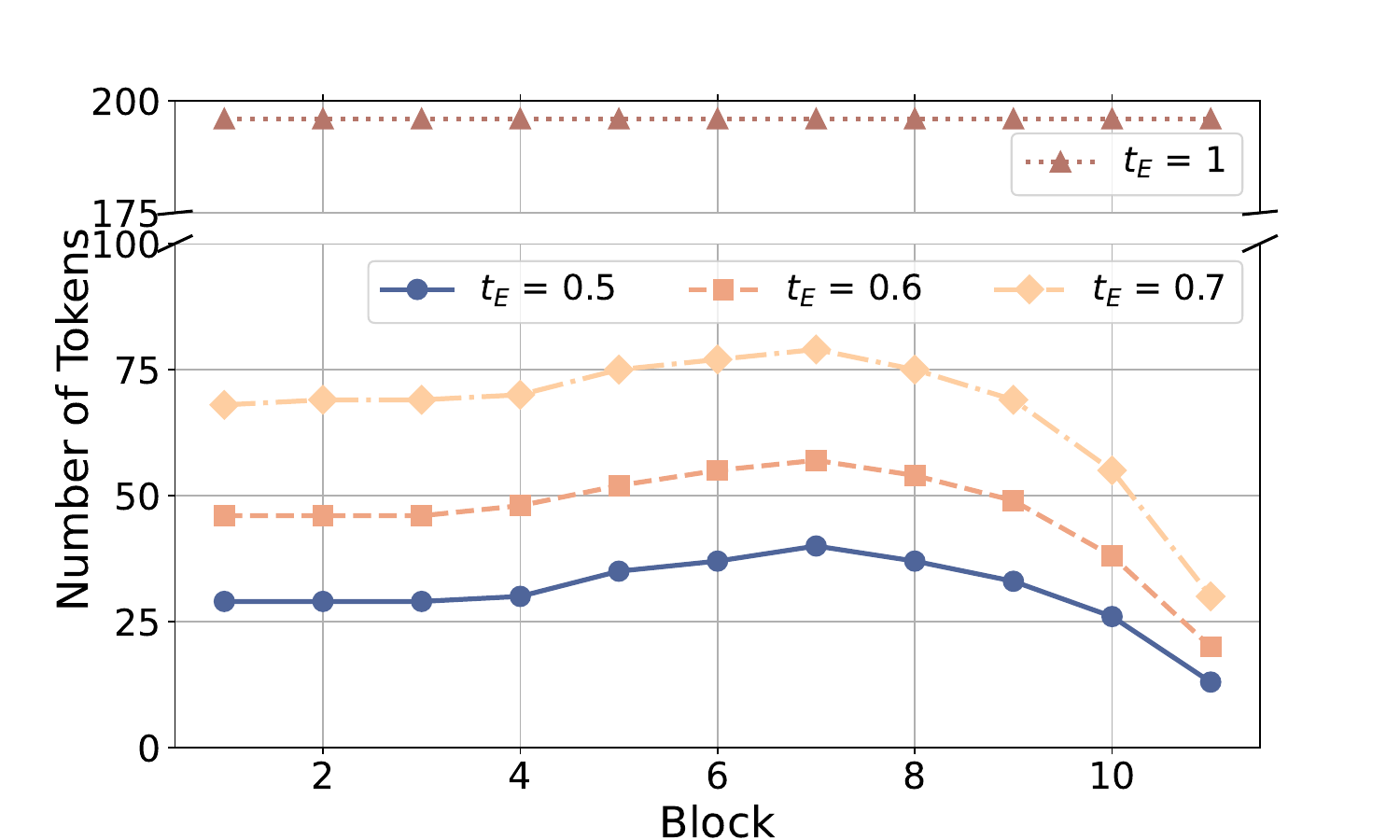}
\vspace{-0.2cm}
\caption{\textbf{Minimum Number of Tokens with Energy $t_{E}$ Across Blocks.} We evaluate on three types of $t_{E}$ for every block.}
\label{fig:token-vs-layer}
\end{figure}

\subsection{Token Reduction Impact}
\label{sec:reduction}
To investigate the intuitive relation between the reduced number of tokens and information drop, we leverage the Singular Value Decomposition (SVD) to monitor the energy of one reduced token sequence. Given a token sequence $\mathbf{X} \in \mathbb{R}^{n \times d}$ (with $n$ being the number of tokens and $d$ their size), we can decompose it applying SVD on $\mathbf{X}^T$:
\begin{equation}
    \mathbf{X}^T = \mathbf{U} \mathbf{\Sigma} \mathbf{V}^T,
\end{equation}
where $\mathbf{U} \in \mathbb{R}^{d \times d}$ and $\mathbf{V} \in \mathbb{R}^{n \times n}$ are two orthogonal matrices and $\mathbf{\Sigma} \in \mathbb{R}^{d \times n}$ is a diagonal matrix, with the singular values $\sigma_i = \mathbf{\Sigma}_{ii}$  representing the variances (energy) in the new compositional directions. In this way, SVD yields the \emph{optimal} energy concentration that can be ordered by their contribution to the total variance (energy)~\cite{eckart1936approximation}. 
Based on this, we can estimate how much variance we preserve by only taking the largest $k$ singular values, as:
\begin{equation}
    E(k) = \frac{\sum_{i=1}^k \sigma_i}{\sum_{i=1}^n \sigma_i},\qquad \sigma_i \geq \sigma_{i+1} \quad \forall i.
\end{equation}

By the above definition, SVD provides an \underline{\emph{upper bound}} on the amount of energy preserved by keeping $k$ tokens. Therefore, by setting a threshold $t_{E}$ of energy to be conserved, we can estimate the theoretical number of tokens $k$ needed for each block as:
\begin{equation}
k = \min \{ k \mid E(k) \ge t_E \}.
\end{equation}

\noindent \textbf{Observations.} Setting a threshold $t_E$, $k$ changes across different blocks. To evaluate the potential token reduction, we use our empirical setup to monitor the change in $k$ (once $t_E$ is defined) in different blocks of the model. As shown in Fig.~\ref{fig:token-vs-layer}, to preserve the same amount of energy, later blocks require significantly fewer tokens than earlier ones. These results offer key indications for developing block-adaptive token reduction strategies, suggesting that the first and last blocks might be suitable candidates.

\subsection{Token Propagation Effect}
\label{sec:propagation}
\begin{figure}[t]
\centering
\includegraphics[width=\columnwidth]{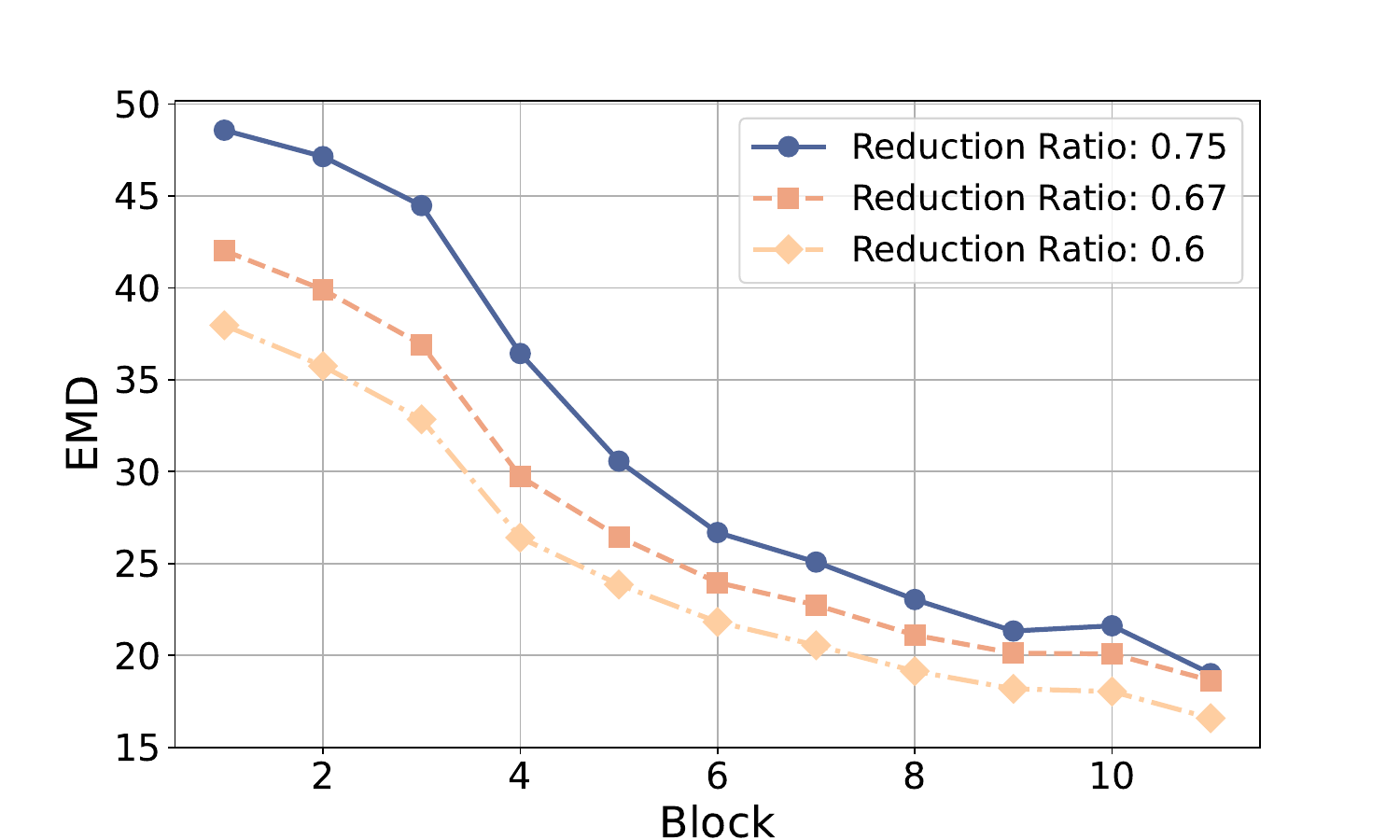}
\caption{\textbf{EMD Distance Between Reduced and Original Output Distributions under 3 Reduction Ratios.} We compare the EMD distance by exerting token reduction on different blocks.} 
\label{fig:EMD-aggregation}
\vspace{-0.1cm}
\end{figure}

While token reduction impact offers an upper bound for information preservation, this approximation does not account for inter-block dependencies and therefore may not adequately reflect the effects on subsequent blocks. Indeed, due to the sequential structure of transformer architectures, errors introduced by token reduction from former blocks can propagate through the network. Similar to the butterfly effects, this impact can accumulate or even amplify through successive modules and non-linear transformations~\cite{block2023butterfly}, thus influencing the final performance.

To analyze this propagation effect, we examine how token reduction at different blocks affects the final output distribution. Thus, by comparing the original output token distribution (without reduction) and the one obtained by operating token reduction only in one block $b$, we can show the presence of such a propagation effect.

To measure this distortion, we consider output tokens from the original model $\mathbf{Y}\in \mathbb{R}^{n \times d}$ and the ones obtained when $n-k$ tokens are reduced in the block $b$: $\mathbf{\tilde{Y}}_b \in \mathbb{R}^{k \times d}$. Then, we define the empirical distributions $P_Y$ and $P_{\tilde{Y}_b}$ supported on $\mathbf{Y}$ and $\mathbf{\tilde{Y}}_b$ respectively, and we evaluate Earth Mover's Distance (EMD)~\cite{rubner2000earth} between them: 
\begin{equation}
    \label{eq:emd}
    \mathrm{EMD}(P_{Y},P_{\tilde{Y}_b}) = \min_{\gamma} \langle \mathbf{\gamma}, \mathbf{M} \rangle_F, 
\end{equation}
where $\mathbf{M} = (d_{ij})_{n \times k} = (\| y_i - \tilde{y}_{bj} \|_2)_{n \times k}$ denotes the metric cost matrix, with $y_i$ and $\tilde{y}_{bj}$ being points in the support of distributions $P_{Y}$ and $P_{\tilde{Y}_b}$ respectively. $\mathbf{\gamma} = (\gamma_{ij})_{n \times k}$ represents the optimal transport matrix, which satisfies the non-negative constraint $\gamma_{ij} \geq 0, \forall i,j$ and the marginal constraints $\sum_{j=1}^k \gamma_{ij} = P_{Y}(y_i)$ and $\sum_{i=1}^n \gamma_{ij} = P_{\tilde{Y}_b} (\tilde{y}_{bj})$.

\begin{figure}[t]
\centering
\includegraphics[width=\columnwidth]{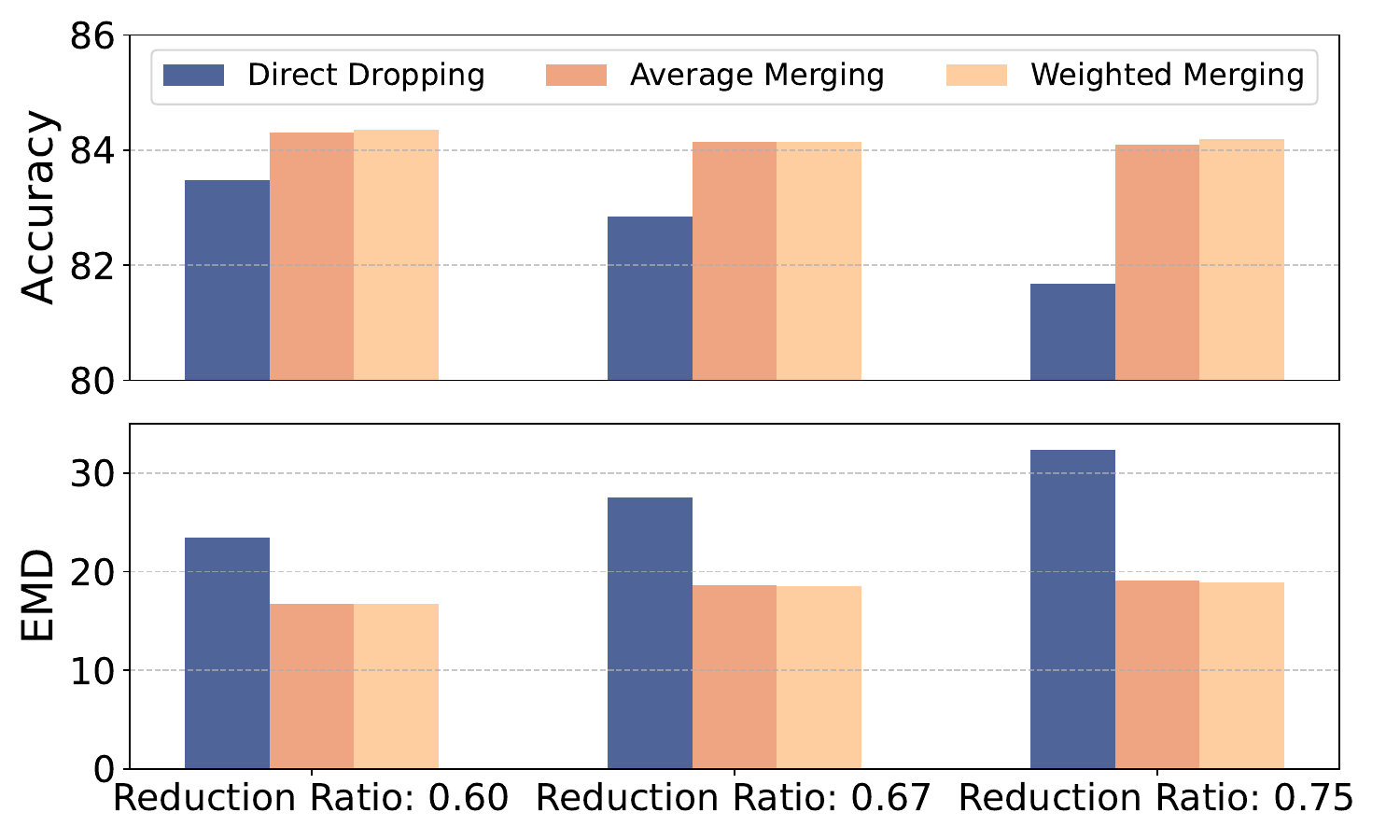}
\vspace{-0.5cm}
\caption{\textbf{Comparison of EMD and Accuracy for Aggregation Methods.} We compare ``direct dropping'', ``average merging'' and ``weighted merging'' under different reduction ratios.} 
\label{fig:aggregation-method}
\end{figure}

\vspace{0.1cm}
\noindent \textbf{Observations.} Due to the propagation effect, the token reduction operation exerting on different blocks can cause distinct levels of distortion on the output distribution. We thus monitor EMD~\eqref{eq:emd} when the same number of tokens is reduced in each specific block. As evidenced in Fig.~\ref{fig:EMD-aggregation}, reducing tokens in early blocks results in substantially higher EMD values compared to later ones. This observation empirically confirms the presence of such propagation effect, where the distortion is amplified across the layers of the network. For this reason, first blocks are finally not suitable for token reduction, different from what is suggested in Fig.~\ref{fig:token-vs-layer}. On the other hand, Fig.~\ref{fig:EMD-aggregation} shows that even with an aggressive reduction ratio of \emph{75\%}, EMD remains remarkably low in the last layers. Synthesizing these findings on reduction impact and propagation effect, we conclude that token reduction should be strategically applied at the end of the network, which allows a high degree of reduction. Results on BLIP~\cite{li2022blip} provided in Tab.~\ref{tab:ablation-blip} confirm these conclusions. More results supporting these observations are provided in supplementary materials.

\begin{figure*}
\centering
\includegraphics[width=\textwidth]{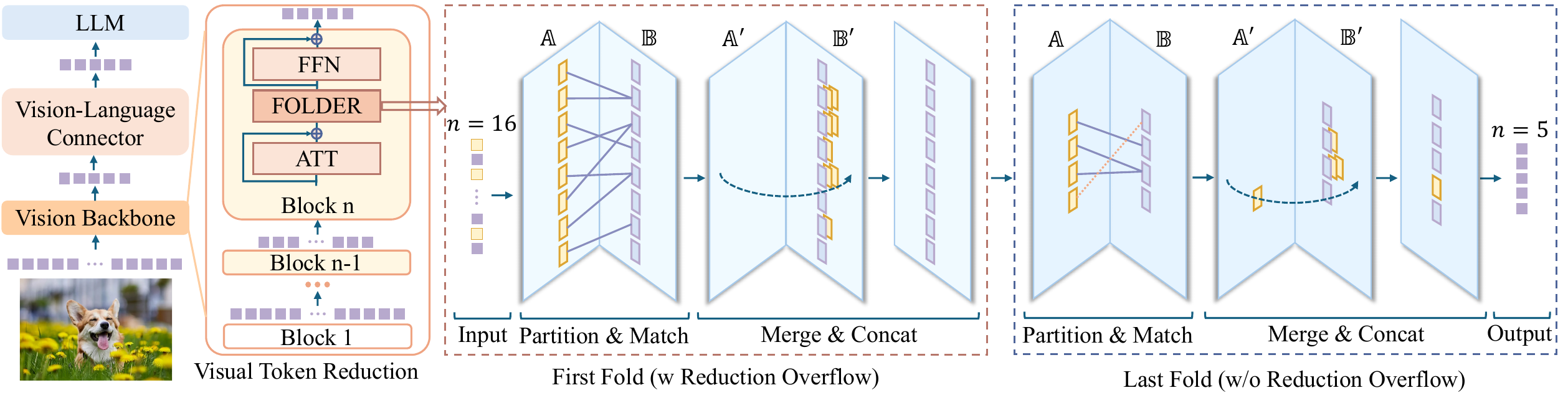}
\vspace{-0.5cm}
\caption{\textbf{Pipeline of FOLDER}. As a plug-and-play module, FOLDER is integrated into the final blocks of the vision backbone (last two here). To deal with reduction overflow, FOLDER automatically executes another FOLD operation when the expected reduction is more than half. The last FOLD, which escapes from reduction overflow, merges tokens according to the remaining reduction numbers.}
\label{fig:pipeline}
\vspace{-0.1cm}
\end{figure*}

\subsection{Token Aggregation Method}
\label{sec:aggregation}
In Sec.~\ref{sec:reduction}, we leverage SVD to compute the upper bound for information preservation. However, due to the computational complexity of SVD, it cannot be directly applied for token reduction. Thus, leveraging what is done in previous works~\cite{bolya2022token,ju2025turbo}, we decompose this operation into two stages: i.) Token Matching: grouping tokens according to a specific matching function, and ii.) Token Aggregation: consolidating the grouped tokens into a reduced number of tokens. While these works~\cite{bolya2022token,ju2025turbo} offered insights on matching functions, the latter stage is barely studied. More formally, given a set of tokens $\{\mathbf{x}_1, \mathbf{x}_2, \ldots, \mathbf{x}_m\}$ that are expected to be aggregated into a new token $\mathbf{y}$, we formulate the aggregation operation as follows:
\begin{equation}
    \mathbf{y} = \sum_{i = 1}^{m} \alpha_{i} \mathbf{x}_i, \text{ s.t. } \sum_{i = 1}^{m} \alpha_{i} = 1.
    \label{eq:15}
\end{equation}
Based on Eq.~\eqref{eq:15}, we formulate three different token aggregation approaches: i.) Average Merging, with $~{\alpha_{i}=\frac{1}{m}, \forall i}$; ii.) Weighted Merging, with $~{\alpha_{i}=\frac{\|x_i\|_2}{\sum_{j = 1}^{m} \|x_j\|_2}, \forall i}$; iii.) Direct Dropping, with $~{\alpha_{i}=0, \forall i \neq i_{max},\alpha_{i_{max}}=1}$ where $i_{max}={\arg} \underset{i}{\max} \|x_i\|_2$. Here the norm $\|\cdot \|_2$ is used as the importance score for tokens.

\vspace{0.1cm}
\noindent\textbf{Observations.}
Similarly with Sec.~\ref{sec:propagation}, we compare these different aggregation methods leveraging the EMD~\eqref{eq:emd}. In Fig.~\ref{fig:aggregation-method}, we notice that with both merging methods we obtain lower discrepancy compared to the Direct Dropping method. Given that Average Merging achieves comparable results to Weighted Merging with greater simplicity, we select it as our preferred aggregation method. Moreover, we can notice that a lower EMD corresponds to higher final accuracy. More results supporting these observations are provided in Tab.~\ref{tab:ablation} and in supplementary materials.

\subsection{FOLDER} \label{sec:folder}
\noindent \textbf{Motivation.} Based on empirical analyses, we deduce that: i.) Token reduction should take place in the last blocks, where aggressive reduction is permitted. ii.) Merging is better than Dropping for aggregation. Therefore, our goal is to design an algorithm that can reduce as many tokens as possible in one block. Previous works on token merging mainly use bipartite matching~\cite{bolya2022token}, which is limited by a maximum reduction ratio of $1/2$. Thus, to aggressively reduce tokens in one block, we propose \emph{FOLDER}, a new bipartite matching strategy that can reduce an arbitrary number of tokens.

\vspace{0.1cm}
\noindent \textbf{Strategy.} As illustrated in Fig.~\ref{fig:pipeline}, our strategy focuses on the last blocks (or the last) in the vision backbone. Here the target reduction number $r$ could exceed half of the total tokens $\left\lfloor n/2 \right\rfloor$ in one block. We refer to this as \textit{reduction overflow}. To handle this case, \emph{FOLDER} automatically applies iteratively a ``FOLD'' operation. As detailed in Alg.~\ref{Alg: FOLDER}, if reduction overflow occurs in a fold iteration (meaning that $r_{\text{remain}} > \left\lfloor n'/2 \right\rfloor$ in line~\ref{algo: folder-overflow}), we reduce the tokens by half ($r_\text{fold} = \left\lfloor n'/2 \right\rfloor$). To do this, we first split the token input sequence into equal-sized partitions named $\mathbb{A}$ and $\mathbb{B}$ (line \ref{algo: folder-split}). Then, for each token in $\mathbb{A}$ we identify the most similar one in set $\mathbb{B}$ based on a Matching Function~\cite{bolya2022token,ju2025turbo} $S: (\mathbb{R}^d, \mathbb{R}^d)\rightarrow \mathbb{R}$ (line~\ref{algo: folder-match}). A visual representation is provided in Fig.~\ref{fig:pipeline}, where the ``Partition \& Match'' operation is highlighted. 
These matches are then sorted based on their matching scores ($\overset{s_{ij}}{\rightarrow}$), and only the top $r_\text{fold}$ matches are preserved (line~\ref{algo: folder-sort}). At this point, by merging these matches in $\mathbf{M}_{top}$ from $\mathbb{A}$ to $\mathbb{B}$, we obtain two new sets: $\mathbb{A'}$ and $\mathbb{B'}$. Namely, $\mathbb{B'}$ updates $\mathbb{B}$ by replacing original tokens with matched tokens aggregated using a specific Aggregation Method (Sec.~\ref{sec:aggregation}).  $\mathbb{A'}$, instead, contains the tokens in $\mathbb{A}$ whose matches were discarded in line~\ref{algo: folder-sort}. Clearly, if the reduction overflow occurs, $r_\text{fold} = \left\lfloor n'/2 \right\rfloor$ and thus $\mathbb{A'} = \emptyset$ (``First Fold'' of Fig.~\ref{fig:pipeline}).
The concatenation between $\mathbb{A'}$ and $\mathbb{B'}$ represents the input token sequence for the next ``FOLD'' iteration, in which $r_\text{remain}$ tokens are going to be removed. This process continues through successive folds (line~\ref{folder:while}) until the overflow is no longer encountered. In this case, $r_{\text{fold}} = r_{\text{remain}}$ (line~\ref{algo: folder-overflow}), and $r_{\text{remain}}$ for the next iteration will be equal to $0$ (line~\ref{algo: folder-remain}). A visual representation of this case is provided in Fig.~\ref{fig:pipeline} in the ``Last FOLD'' iteration.

\begin{algorithm}[ht]
\caption{\emph{FOLDER} applied in one block}
\label{Alg: FOLDER}

\KwIn{Token sequence $\mathbf{X}\in \mathbb{R}^{n \times d}$, reduced num $r$}
\KwOut{Reduced token sequence $\mathbf{X}'\in \mathbb{R}^{(n-r) \times d}$}
\SetKwProg{Fn}{Function}{:}{}
\SetKwFunction{FFOLD}{FOLD}
\SetKwFunction{Match}{Macthing}

$r_{\text{remain}} = r$;

$\mathbf{X}' = \mathbf{X}$;

\While{$r_{\text{remain}} > 0$}{ \label{folder:while}
$\mathbf{X}'$, $r_{\text{remain}}$ = \FFOLD $(\mathbf{X}', r_{\text{remain}})$;
}

\rule{.8\columnwidth}{0.4pt}

    \Fn{\FFOLD{$\mathbf{X}'\in \mathbb{R}^{n' \times d}$, $r_{\text{remain}}$, }}{

        $r_{\text{fold}} = \min(\left\lfloor n/2 \right\rfloor
, r_{\text{remain}})$; \label{algo: folder-overflow}

        $r_{\text{remain}} = r_{\text{remain}} - r_{\text{fold}}$;\label{algo: folder-remain}
        
        $\{\mathbb{A}, \mathbb{B} \}$ $\Leftarrow$ Split $\textbf{X}'$ into 2 equal-sized partitions; \label{algo: folder-split}
        
        $\mathbf{M}$ $\Leftarrow$ Compute $\{ (a_i \overset{s_{ij}}{\rightarrow} b_j) \mid b_j=~\underset{b_j \in \mathbb{B}}{\arg\max}~s_{ij} =~S(a_i, b_j), \forall a_i \in \mathbb{A} \}$; \label{algo: folder-match}

        $\mathbf{M}_{top}$ $\Leftarrow$ Sort $\mathbf{M}$ based on $\overset{s_{ij}}{\rightarrow}$ \& keep Top $r_{\text{fold}}$; \label{algo: folder-sort}

        $\{\mathbb{A}', \mathbb{B}' \}$ $\Leftarrow$ Merge $\mathbb{A}$ to $\mathbb{B}$ according to $\mathbf{M}_{top}$; \label{algo: folder-merge}
        
        \Return Concat($\mathbb{A}'$, $\mathbb{B}'$), $r_{\text{remain}}$; \label{algo: folder-return}
}
\end{algorithm}

\begin{table*}[t]
\centering
\footnotesize
\caption{\textbf{Results on LLaVA1.5 7B/13B.} We highlight the best result for each benchmark under the same reduction ratio. $^{*}$Retrained to provide a fair comparison in Tab.~\ref{tab:llava-retrain}. }
\vspace{-0.3cm}
\begin{tabular}{c|c|c|C{0.8cm}C{0.6cm}cC{0.8cm}ccccC{0.8cm}cc}
\hline
Method & \begin{tabular}[c]{@{}c@{}}Reduct \\ Ratio\end{tabular} & \begin{tabular}[c]{@{}c@{}}Speed \\ Up\end{tabular} & \begin{tabular}[c]{@{}c@{}}MMBench\\ EN\end{tabular} & \begin{tabular}[c]{@{}c@{}}MM\\ Star\end{tabular} & MME & \begin{tabular}[c]{@{}c@{}}A-OK\\ VQA\end{tabular} & \begin{tabular}[c]{@{}c@{}}Hallusion\\ Bench\end{tabular} & \begin{tabular}[c]{@{}c@{}}MM\\ MU\end{tabular} & \begin{tabular}[c]{@{}c@{}}OCR\\ VQA\end{tabular} & \begin{tabular}[c]{@{}c@{}}\scriptsize SEED\\ \scriptsize BenchIMG\end{tabular} & \begin{tabular}[c]{@{}c@{}}Science\\ QA\end{tabular} & POPE & Avg \\ \hline  \hline

Original-7B & 0\% & 1$\times$ & 62.8 & 32.7 & 1338.9 & 78.8 & 35.6 & 32.2 & 52.4 & 60.2 & 68.1 & 79.7 & 55.0 \\ \hline
Pooling & \multirow{4}{*}{50\%} & 1.5$\times$ & 59.5 & 30.3 & 1308.6 & 77.1 & 35.3 & 31.7 & 44.8 & 59.2 & 68.0 & 84.0 & 53.7 \\

FastV~\cite{chen2024image} &  & 1.3$\times$ & \textbf{63.3} & \textbf{32.4} & \textbf{1345.1} & \textbf{78.6} & 36.5 & 28.7 & \textbf{53.1} & 59.4 & 67.8 & 81.0 & 54.9 \\

Turbo~\cite{ju2025turbo} &  & 1.5$\times$ & 60.4 & 30.9 & 1311.2 & 77.2 & 35.5 & 28.0 & 35.7 & 58.1 & 67.5 & 83.0 & 52.3\\

Ours & & 1.5$\times$ & 62.4 & 32.1 & 1338.2 & 78.3 & \textbf{38.3} & \textbf{34.0} & 48.5 & \textbf{59.5} & \textbf{68.5} & \textbf{85.4} & \textbf{55.5}  \\ \hline

Pooling & \multirow{4}{*}{66\%} & 1.7$\times$ & 57.6 & 29.7 & 1308.2 & 74.2 & 32.3 & 30.7 & 38.7 & 57.6 & 67.3 & 80.7 & 51.6\\

FastV~\cite{chen2024image} & &  1.5$\times$ & \textbf{62.5} & \textbf{32.0} & \textbf{1353.2} & 77.7 & 37.4 & 30.0 & \textbf{52.6} & 58.3 & 68.2 & 79.3 & 54.6\\

Turbo~\cite{ju2025turbo} & & 1.7$\times$ & 60.1 & 31.5 & 1301.6 & \textbf{78.0} & 34.9 & 24.7 & 34.4 & 57.8 & 67.2 & 85.0 & 52.0 \\

Ours & &  1.7$\times$ & 61.4 & 30.3 & 1350.0 & 77.9 & \textbf{39.2} & \textbf{31.3} & 46.1 & \textbf{59.7} & \textbf{68.3} & \textbf{85.4} & \textbf{54.8} \\
\hline
FastV~\cite{chen2024image} & \multirow{2}{*}{75\%} & 1.6$\times$ & 61.2 & 31.0 & 1321.6 & 76.6 & 37.5 & 32.0 & \textbf{50.8} & 57.2 & 68.1 & 77.7 & 53.9\\ 

Ours+FastV & & 1.7$\times$ & \textbf{61.4} & \textbf{31.2} & \textbf{1325.5} & \textbf{78.0} & \textbf{39.1} & \textbf{32.6} & 48.8 & \textbf{59.2} & \textbf{68.6} & \textbf{82.3} & \textbf{54.9}\\ 
\hline \hline
Original-13B$^{*}$ & 0\% & 1$\times$ & 66.6 & 30.9 & 1371.1 & 77.0 & 36.1 & 34.0 & 54.9 & 59.4 & 68.8 & 86.4 & 56.3\\ \hline
Pooling & \multirow{4}{*}{50\%} & 1.5$\times$ & 64.1 & 31.2 & 1316.6 & 76.2 & 35.1 & 30.0 & 54.7 & 57.6 & 69.0 & 85.7 & 55.1 \\

FastV~\cite{chen2024image} &  & 1.3$\times$ & \textbf{66.4} & 31.1 & \textbf{1386.3} & 75.5 & 36.2 & 32.6 & 54.9 & 58.9 & 68.4 & 85.4 & 55.9\\

Turbo~\cite{ju2025turbo} &  & 1.5$\times$ & 65.0 & 30.6 & 1200.2 & 78.2 & 26.9 & 32.7 & 47.9 & 58.8 & 69.7 & 86.1 & 53.9 \\

Ours &  & 1.5$\times$ & 65.4 & \textbf{31.2} & 1383.7 & \textbf{78.6} & \textbf{36.4} & \textbf{34.4} & \textbf{56.2} & \textbf{59.1} & \textbf{70.5} & \textbf{86.9} & \textbf{56.8} \\ \hline
Pooling & \multirow{4}{*}{66\%} & 1.6$\times$ & 62.8 & 31.0 & 1250.6 & 73.5 & 29.5 & 32.5 & 45.6 & 54.5 & 69.7 & 82.1 & 52.6\\

FastV~\cite{chen2024image} & & 1.5$\times$ & \textbf{66.3} & 31.2 & 1352.2 & 75.0 & \textbf{35.5} & 32.3 & \textbf{54.4} & 58.0 & 68.3 & 83.5 & 55.3 \\

Turbo~\cite{ju2025turbo} &  & 1.6$\times$ & 64.7 & 31.2 & 1168.3 & 76.3 & 25.9 & 33.3 & 47.9 & 58.5 & 70.5 & 85.5 & 53.6\\
Ours & & 1.6$\times$ & 65.8 & \textbf{31.7} & \textbf{1366.9} & \textbf{77.3} & 33.8 & \textbf{35.0} & 52.6 & \textbf{58.8} & \textbf{70.7} & \textbf{86.1} & \textbf{56.1} \\ \hline
FastV~\cite{chen2024image} & \multirow{2}{*}{75\%} & 1.6$\times$ & 64.2 & 31.7 & 1321.3 & 74.2 & \textbf{36.9} & 31.3 & 53.3 & 56.4 & 68.1 & 82.1 & 54.5 \\
Ours+FastV & & 1.8$\times$ & \textbf{65.1} & \textbf{32.3} & \textbf{1368.6} & \textbf{77.8} & 35.3 & \textbf{32.0} & \textbf{55.2} & \textbf{58.8} & \textbf{71.1} & \textbf{85.8} & \textbf{56.2} \\ 
\hline
\end{tabular}
\label{tab:llava}
\end{table*}

\begin{table*}[t]
\centering
\footnotesize
\caption{\textbf{Results on Minigpt4v2.} We highlight the best result on each benchmark under the same reduction ratio.}
\vspace{-0.3cm}
\begin{tabular}{c|c|c|C{0.9cm}C{0.9cm}C{0.8cm}C{0.8cm} cccC{1.1cm}C{0.8cm}C{0.8cm}C{0.8cm}}
\hline
Method & \begin{tabular}[c]{@{}c@{}}Reduct \\ Ratio\end{tabular} & \begin{tabular}[c]{@{}c@{}}Speed \\ Up\end {tabular} & \begin{tabular}[c]{@{}c@{}} MMBench\\ EN\end{tabular} & \begin{tabular}[c]{@{}c@{}}MM\\ Star\end{tabular} & MME & \begin{tabular}[c]{@{}c@{}}A-OK\\ VQA\end{tabular} & \begin{tabular}[c]{@{}c@{}}Hallusion\\ Bench\end{tabular} & \begin{tabular}[c]{@{}c@{}}MM\\ MU\end{tabular} & POPE & \begin{tabular}[c]{@{}c@{}}SEED\\ BenchIMG\end{tabular} & \begin{tabular}[c]{@{}c@{}} Science\\ QA\end{tabular} & \begin{tabular}[c]{@{}c@{}}RealW\\ QA\end{tabular} & Avg \\ \hline  \hline
Original & 0\% & 1$\times$ & 9.2 & 24.4 & 631.4 & 38.5 & 26.4 & 18.7 & 62.3 & 31.8 & 48.6 & 38.8 & 32.1\\ \hline

Turbo & \multirow{2}{*}{50\%} & 1.3$\times$ & 8.7 & 21.0 & 692.1 & 35.0 & \textbf{34.7} & 20.7 & 52.7 & 30.5 & 50.4 & 37.6 & 31.6 \\

Ours & & 1.3$\times$ & \textbf{9.4} & \textbf{22.5} & \textbf{710.3} & \textbf{38.4} & 34.6 & \textbf{24.0} & \textbf{56.2} & \textbf{31.3} & \textbf{50.7} & \textbf{38.6} & \textbf{33.1}\\ \hline

Turbo & \multirow{2}{*}{60\%} & 1.4$\times$ & 9.1 & 22.1 & 674.5 & 36.7 & 34.3 & 23.0 & 51.0 & 30.4 & 50.3 & 38.8 & 32.0\\

Ours & & 1.4$\times$ & \textbf{13.8} & \textbf{24.3} & \textbf{859.9} & \textbf{43.8} & \textbf{35.6} & \textbf{23.1} & \textbf{63.3} & \textbf{33.7} & \textbf{53.5} & \textbf{39.9} & \textbf{36.2}\\ \hline

Turbo & \multirow{2}{*}{70\%} & 1.6$\times$ & 8.7 & 22.0 & 651.8 & 36.3 & 34.4 & 20.7 & 44.7 & 30.7 & 50.1 & 35.0 & 30.6 \\

Ours & & 1.6$\times$ & \textbf{9.3} & \textbf{22.5} & \textbf{666.8} & \textbf{37.6} & \textbf{35.5} & \textbf{22.8} & \textbf{56.6} & \textbf{31.2} & \textbf{51.5} & \textbf{37.9} & \textbf{32.9} \\ \hline
\end{tabular}
\label{tab:minigpt}
\end{table*}

\begin{table*}[t]
\centering
\footnotesize
\caption{\textbf{Results on MMVP (MLLM with Multiple Vision-towers (CLIP+DINOv2).}}
\vspace{-0.3cm}
\begin{tabular}{c|c|c|C{1cm}C{0.6cm}cC{0.8cm}ccccC{0.8cm}cc}
\hline
Method & \begin{tabular}[c]{@{}c@{}}Reduct \\ Ratio\end{tabular} & \begin{tabular}[c]{@{}c@{}}Speed \\ Up\end {tabular} & \begin{tabular}[c]{@{}c@{}} MMBench\\ EN\end{tabular} & \begin{tabular}[c]{@{}c@{}}MM\\ Star\end{tabular} & MME & \begin{tabular}[c]{@{}c@{}}Hallusion\\ Bench\end{tabular} & \begin{tabular}[c]{@{}c@{}}OCR\\ VQA\end{tabular} & \begin{tabular}[c]{@{}c@{}}MM\\ MU\end{tabular} & POPE & \begin{tabular}[c]{@{}c@{}}SEED\\ BenchIMG\end{tabular} & \begin{tabular}[c]{@{}c@{}}Science\\ QA\end{tabular} & \begin{tabular}[c]{@{}c@{}}CC\\ Bench\end{tabular} & Avg \\ \hline  \hline
Original & 0\% & 1$\times$ & 64.2 & 32.1 & 1403.6 & 46.3 & 53.3 & 34.2 & 86.7 & 60.6 & 70.0 & 18.8 & 51.6 \\ \hline

Turbo~\cite{ju2025turbo} & \multirow{2}{*}{50\%} & 1.5$\times$ & 62.8 & 32.0 & 1352.9 & 44.0 & 47.6 & 34.0 & 85.6 & 58.6 & 69.1 & 17.9 & 50.0 \\

Ours & & 1.5$\times$ &\textbf{63.0} & \textbf{32.1} & \textbf{1362.3} & \textbf{46.2} & \textbf{50.5} & \textbf{34.7} & \textbf{85.7} & \textbf{59.5} & \textbf{70.4} & \textbf{18.6} & \textbf{50.9} \\ \hline

Turbo~\cite{ju2025turbo} & \multirow{2}{*}{66\%} & 1.7$\times$ & 62.5 & 30.7 & 1341.8 & 46.2 & 46.6 & 33.3 & 84.9 & 58.1 & 69.7 & 18.8 & 49.9 \\

Ours & & 1.7$\times$ & \textbf{63.6} & \textbf{31.9} & \textbf{1368.2} & \textbf{47.9} & \textbf{47.2} & \textbf{34.0} & \textbf{85.3} & \textbf{58.7} & \textbf{70.7} & \textbf{19.6} & \textbf{50.8} \\ \hline
\end{tabular}
\label{tab:mmvp}
\end{table*}

\begin{table*}[t]
\centering
\footnotesize
\caption{\textbf{Training with FOLDER.} w/ F: We plug FOLDER into LLaVA1.5-13B during training ($\scriptstyle \text{Train Reduct}>0$) and inference ($\scriptstyle \text{Infer Reduct}>0$). Training with FOLDER yields an all-round enhancement on both performance and speed.}
\vspace{-0.3cm}
\begin{tabular}{c|c|c|c|C{1cm}C{0.6cm}cC{0.8cm}ccccC{0.8cm}cc}
\hline
Train & \begin{tabular}[c]{@{}c@{}}Train \\ Reduct \end{tabular} &\begin{tabular}[c]{@{}c@{}}Infer \\ Reduct\end{tabular} &\begin{tabular}[c]{@{}c@{}}Speed \\ Up\end{tabular} & \begin{tabular}[c]{@{}c@{}}MMBench\\ EN\end{tabular} & \begin{tabular}[c]{@{}c@{}}MM\\ Star\end{tabular} & MME & \begin{tabular}[c]{@{}c@{}}A-OK\\ VQA\end{tabular} & \begin{tabular}[c]{@{}c@{}}Hallusion\\ Bench\end{tabular} & \begin{tabular}[c]{@{}c@{}}OCR\\ VQA\end{tabular} & \begin{tabular}[c]{@{}c@{}}SEED\\ BenchIMG\end{tabular} & \begin{tabular}[c]{@{}c@{}}Science\\ QA\end{tabular} & POPE & Avg \\ \hline  \hline
w/o F & 0\% & 0\% & (1,1)$\times$ & 66.6 & 30.9 & 1371.1 & 77.0 & 36.1 & 54.9 & 59.4 & 68.8 & 86.4 & 58.8 \\ \hline
w/ F & 0\% & 66\% & (1,1.6)$\times$ & 65.8 & 31.7 & 1366.9 & 77.3 & 33.8 & 57.6 & 58.8 & 70.7 & 86.1 & 59.0 \\
w/ F & 66\% & 0\% & (1.5,1)$\times$ & 66.6 & 32.0 & \textbf{1407.7} & 78.9 & 42.8 & 57.7 & 60.9 & 69.3 & \textbf{88.3} & 60.8 \\
w/ F & 66\% & 66\% & (1.5,1.6)$\times$ & \textbf{67.8} & \textbf{32.5} & 1393.4 & \textbf{79.5} & \textbf{43.6} & \textbf{57.9} & \textbf{61.3} & \textbf{71.5} & 86.9 & \textbf{61.2} \\ \hline
w/ F & 75\% & 0\% & (1.7,1)$\times$ & 63.3 & 30.7 & 1359.5 & 79.1 & 37.7 & 56.2 & 60.2 & 65.6 & 87.8 & 58.8 \\
w/ F & 75\% & 75\% & (1.7,1.8)$\times$ & 65.2 & 30.7 & 1325.7 & 79.5 & 34.4 & 52.3 & 59.5 & 67.4 & 87.2 & 58.2 \\ \hline
\end{tabular}
\label{tab:llava-retrain}
\vspace{-0.2cm}
\end{table*}

\section{Experiments}
In this section, we present the results. By using FOLDER, we reduce the number of visual tokens to accelerate both training and inference of MLLMs.

\subsection{Evaluation \& Benchmarks}
\noindent \textbf{Evaluation Models.} 
We evaluate our method on various MLLMs, including image understanding models LLaVA1.5-7B/13B~\cite{liu2024improved} and Minigpt4v2~\cite{chen2023minigpt}, multi-vision-tower based model MMVP~\cite{tong2024eyes} and video understanding model Video-LLaVA~\cite{lin2023video}. For merging strategy ablation, we evaluate on BLIP~\cite{li2022blip} for the image captioning task.

\vspace{0.1cm}
\noindent \textbf{Benchmarks.}
To quantify the ability of accelerated MLLMs, we conduct thorough experiments on a wide range of tasks, including \underline{\textit{Y/N tasks}} like MME~\cite{fu2023mme}, HallusionBench~\cite{guan2024hallusionbench}, POPE~\cite{Li-hallucination-2023}; \underline{\textit{MCQ tasks}} like MMBench-EN~\cite{MMBench}, CCBench~\cite{MMBench}, A-OKVQA~\cite{schwenk2022okvqa}, ScienceQA\_IMG~\cite{lu2022learn}, SEEDBench\_IMG~\cite{li2023seed}, MMMU~\cite{yue2023mmmu}, MMStar~\cite{chen2024we}, Video-MME~\cite{fu2024video}, RealWorld-QA~\cite{grok}; and \underline{\textit{VQA tasks}} like OCRVQA~\cite{mishraICDAR19}, MMBench-Video~\cite{fang2024mmbenchvideo}. For BLIP, we evaluate on COCO image captioning~\cite{lin2014microsoft}. Note that due to cost budget, we only apply LLM evaluation on VQA tasks (exact matching for the others).

\vspace{0.1cm}
\noindent \textbf{Retraining.}
To validate the effectiveness of FOLDER in MLLM training, we leverage LLaVA1.5-13B~\cite{liu2024improved}. We adopt the same training setting of LLaVA1.5-13B, and apply FOLDER during both pretraining and SFT. Training details are provided in the supplementary material.

\subsection{Main Results}
The following experiments reveal the effectiveness of FOLDER on different MLLMs. If not specified, we only apply FOLDER in the last layer of the visual encoder.

\subsubsection{Inference Acceleration on MLLM}
\label{sec:inference_folder}
We evaluate FOLDER's plug-and-play acceleration performance in inference phase on various image understanding MLLMs, including LLaVA1.5-7B, LLaVA1.5-13B~\cite{liu2024improved}, Minigpt4v2~\cite{chen2023minigpt} and multi-vision-tower based model MMVP~\cite{tong2024eyes}. If not specified, we evaluate MLLM's speed-up based on the time of the first output token.

\vspace{0.1cm}
\noindent \textbf{Single-Vision-Tower MLLM.} 
In Tab.~\ref{tab:llava}, we compare FOLDER with previous plug-and-play SOTA~\cite{chen2024image,ju2025turbo} on LLaVA1.5. Overall, FOLDER achieves the best performance-speed trade-off, comparable to or even exceeding the original performance with over 50\% reduction on several benchmarks. Compared to Turbo~\cite{ju2025turbo}, which employs uniform merging across the vision encoder, our approach demonstrates superior performance across the majority of benchmarks under the same reduction ratio. This advantage is particularly pronounced in visually dense tasks such as OCRVQA, where our method significantly outperforms Turbo. This result well accords with our preliminary observation: instead of progressive reduction, tokens should be reduced only in the last blocks.

Despite FastV~\cite{chen2024image} demonstrating comparable performance with FOLDER on some benchmarks, due to its limitations described in Sec.~\ref{sec:related work}, it's inherently slower and does not provide acceleration support for training. FOLDER, on the other hand, is a a ready-to-use plug-and-play module that integrates seamlessly into the vision backbone with minimal adjustment. By directly reducing visual tokens before interfacing with the LLM, FOLDER effectively mitigates issues associated with complex LLM architectures. Note that FOLDER and FastV address token reduction from two distinct perspectives (visual redundancy and task-oriented attention). Thus, as shown in Tab.~\ref{tab:llava}, coupling FOLDER with FastV (50\% $\times$ 50\%) can achieve additional reduction while maintaining competitive performance.

In Tab.~\ref{tab:minigpt}, we also evaluate FOLDER on another MLLM: Minigpt4v2. With a reduction ratio up to 60\%, our method achieves an overall performance improvement compared to the original model, with an increase of more than 40\% on MME and MMBench. Even in this case, by merging in the last block, FOLDER outperforms Turbo. This implies that, in addition to speed enhancement, FOLDER can potentially be a plug-and-play performance booster, reducing the noise occurring in long token sequences.

To prove the effectiveness of our algorithm, we also replaced the matching function~\cite{ju2025turbo} with a naive mean pooling operation between two partitions. The result in Tab.~\ref{tab:llava} demonstrates the superiority of matching over pooling.

\vspace{0.1cm}
\noindent \textbf{Multi-Vision-Tower MLLM.}
Multi-vision-tower-based models~\cite{shi2024eagle,kar2024brave,tong2024eyes} enhance MLLMs' visual ability but aggravate the length issue. This makes our method particularly advantageous in this context. In Tab.~\ref{tab:mmvp} we evaluate on MMVP~\cite{tong2024eyes} with CLIP~\cite{radford2021learning} and DINOv2~\cite{oquab2023dinov2} as vision towers. By applying FOLDER, we trade a significant reduction in image token length (from 1152 to 576/384) with an acceptable drop. Note that MMVP is trained by interleaving vision tower features, which may cause positional confusion. Despite that, FOLDER still achieves a satisfying acceleration-performance trade-off.

\begin{table}[t]
\centering
\footnotesize
\caption{\textbf{Results on VideoLLaVA.}}
\vspace{-0.2cm}
\begin{tabular}{c|c|c|ccc}
\hline
Method & \begin{tabular}[c]{@{}c@{}}Reduct \\ Ratio\end{tabular} & \begin{tabular}[c]{@{}c@{}}Speed \\ Up\end{tabular} & \begin{tabular}[c]{@{}c@{}}MMBench-V\\ Overall\end{tabular} & \begin{tabular}[c]{@{}c@{}}MMBench-V\\Perception \end{tabular} & \begin{tabular}[c]{@{}c@{}}MME\\ Video\end{tabular} \\ \hline  \hline
Original & 0\% & 1$\times$ & 1.04 & 1.04 & 31.8 \\ \hline
Turbo& \multirow{2}{*}{65\%} & 2.5$\times$ & 1.00 & 0.99 & 29.8 \\
Ours &  & 2.4$\times$ & \textbf{1.04} & \textbf{1.05} & \textbf{31.9} \\ \hline
Turbo & \multirow{2}{*}{75\%} & 2.8$\times$ & 0.98 & 0.98 & 29.0  \\
Ours &  & 2.7$\times$ & \textbf{1.03} & \textbf{1.04} & \textbf{30.7} \\ \hline
\end{tabular}
\label{tab:videollava}
\vspace{-0.1cm}
\end{table}

\vspace{0.1cm}
\noindent \textbf{Video Understanding MLLM.} 
Video-based MLLMs~\cite{li2023videochat,lin2023video,jin2024chat} can generate sequences exceeding 2000 tokens for a few frames, limiting their applicability. In Tab.~\ref{tab:videollava}, we show that, by applying FOLDER on Video-LLaVA, we can reduce visual tokens up to 65\%, without any performance degradation. These sequences can be further reduced up to 75\%, with a very slight \& acceptable drop on some benchmarks. Note that we speed up the inference by 2.7$\times$ by reducing 75\% visual tokens, and the result is still much better than Turbo with 66\% reduction ratio.

\begin{table}[t]
\centering
\footnotesize
\caption{\textbf{Acceleration For Training \& Inference.}We evaluate training (A100-80G GPU hours) and inference acceleration (on 1st-token) on llava1.5-13B. The memory is evaluated by its peak value on H20 with batch-size 40. Note that FastV~\cite{chen2024image} can only be used during inference.}
\vspace{-0.2cm}
\begin{tabular}{c|c|cc|ccc}
\toprule
\multirow{2}{*}{Method} & \begin{tabular}[c]{@{}c@{}}\footnotesize Reduct \\ \footnotesize Ratio\end{tabular} & \multicolumn{2}{c|}{Train} & \multicolumn{3}{c}{Inference} \\
&  & Time(h) & \begin{tabular}[c]{@{}c@{}}Speed \\ Up\end{tabular} & Time(s) & \begin{tabular}[c]{@{}c@{}}Speed \\ Up\end{tabular} & Mem \\
\hline \hline
Original & 0\% & 187 & 1$\times$ & 0.365 & 1$\times$ & 81.4G \\ \hline
FastV & \multirow{3}{*}{50\%} & -- &  -- & 0.275& 1.3$\times$ & 81.4G \\
Turbo &  & 139 & 1.34$\times$ & 0.250 & 1.5$\times$ & 56.6G \\
Ours &  & 141 & 1.33$\times$ & 0.250 & 1.5$\times$ & 56.6G \\ \hline
FastV & \multirow{3}{*}{66\%} & -- & -- & 0.255 & 1.4$\times$ & 81.4G \\
Turbo &  & 123 & 1.52$\times$ & 0.230 & 1.6$\times$ & 48.6G \\
Ours &  & 124 & 1.51$\times$ & 0.230 & 1.6$\times$ & 48.7G \\ \hline
FastV & \multirow{3}{*}{75\%} & -- & -- & 0.225 & 1.6$\times$ & 81.4G \\
Turbo &  & 112 & 1.67$\times$ & 0.205 & 1.8$\times$ & 44.6G \\
Ours &  & 113 & 1.65$\times$ & 0.205 & 1.8$\times$ & 44.7G \\
\bottomrule
\end{tabular}
\label{tab:throughput}
\vspace{-0.2cm}
\end{table}

\subsubsection{Training Acceleration on MLLM}
\label{sec:train_folder}
\begin{table}[t]
\centering
\footnotesize
\caption{\textbf{Ablation Study on Merging Position.} For BLIP~\cite{li2022blip}, we evaluate different merging positions on layers under the image captioning task. Last-$n$ refers to uniform reduction in last $n$ blocks, uniform refers to uniform reduction in every block.}
\vspace{-0.3cm}
\begin{tabular}{c|c|c|c|cc}
\toprule
Method & \begin{tabular}[c]{@{}c@{}}Reduct \\ Ratio\end{tabular} &  \begin{tabular}[c]{@{}c@{}}Throughput \\im/s\end{tabular} & \begin{tabular}[c]{@{}c@{}}Speed \\ Up\end{tabular} & CIDEr & B@4 \\
\hline \hline
Original & 0\% & 22.1 & 1$\times$ & 133.3 & 39.7 \\ \hline
Uniform & \multirow{4}{*}{60\%} & 40.2 & 1.82$\times$ & 128.8 & 37.9 \\
Last-3 & & 38.4 & 1.74$\times$ & 129.5 & 38.2 \\
Last-2 & & 37.9 & 1.71$\times$ & 129.4 & 38.2 \\
Last-1 & & 37.5 & 1.71$\times$ & \textbf{132.2} & \textbf{39.0} \\
\hline
Uniform & \multirow{4}{*}{70\%} & 47.4 & 2.14$\times$ & 127.2 & 37.5 \\
Last-3 &  & 44.2 & 2.00$\times$ & 128.4 & 38.2 \\
Last-2 &  & 43.7 & 1.98$\times$ & 130.1 & 38.4 \\
Last-1 & & 43.4 & 1.96$\times$ & \textbf{131.0} & \textbf{38.9}  \\
\bottomrule
\end{tabular}
\label{tab:ablation-blip}
\vspace{-0.1cm}
\end{table}

As an universal plug-and-play module for MLLM, FOLDER can be seamlessly integrated into MLLM pretraining and SFT. We conduct such an experiment on LLaVA1.5-13B, providing results in Tab.~\ref{tab:llava-retrain}. Surprisingly, when incorporating FOLDER with a 66\% reduction ratio, LLaVA-1.5-13B exhibits comprehensive improvements across all benchmarks, even including visually dense tasks such as OCRVQA ($\uparrow$3\%) and visually demanding evaluations like the Hallusion Bench ($\uparrow$7.5\%). These results suggest the presence of potential low-frequency noise within the visual sequence. By merging similar visual tokens, we effectively smooth this low-frequency noise, thereby enhancing the model's learning process. Moreover, as noted in~\cite{chen2024image}, attention sparsity on visual tokens is observed in the LLM attention layers, highlighting the need to reduce visual sequence length to make the model more ``focused''. 
Furthermore, training on a reduced sequence does not compromise flexibility during inference; the original sequence can still be used at inference time, yielding improved results on certain tasks. Similarly to masked augmentation~\cite{he2022masked}, FOLDER acts as a regularization mechanism. However, unlike random dropping, which is limited to the training phase, FOLDER can leverage the same aggregation framework to accelerate both training and inference.

\vspace{0.1cm}
\noindent \textbf{Evaluation for Speed \& Memory.}
On LLaVA1.5-13B, Tab.~\ref{tab:throughput} tests FOLDER's acceleration during training and inference . We train the model on 8 A100-80G and do inference on H20-96G. By reducing 66\% of the tokens, we speed up the training by 1.5$\times$ and enjoy 1.6$\times$ acceleration during inference. Additionally, unlike FastV, which prunes visual tokens within the LLM, FOLDER reduces the visual tokens before the LLM, thereby significantly lowering peak memory consumption. In tests conducted with a batch size of 40 on an H20-96G, FOLDER achieves a 40\% memory savings at a 66\% reduction ratio. This method ideally allows for an increase in maximum batch size which amounts to 3$\times$ for a 66\% reduction-particularly advantageous for training.

\subsection{Ablation Study} \label{subsec: ablation}
We conduct ablation studies on BLIP~\cite{li2022blip} for image captioning task, to illustrate the idea of merging position and aggregation method discussed in Sec.~\ref{sec:method}.

\vspace{0.1cm}
\noindent \textbf{Merging Position.}
As discussed in Sec.~\ref{sec:propagation}, we conduct experiments on the merging position, namely in which block we should merge the tokens. According to preliminary tests on ViT (Fig.~\ref{fig:token-vs-layer}, Fig.~\ref{fig:EMD-aggregation}), we find out that due to the propagation effect and the decreasing trend of reduction impact, we should reduce more tokens in the last blocks rather than early blocks. Following this observation, we choose 4 different merging positions: uniform merging in each block and uniform merging in the last 1/2/3 blocks. In Tab.~\ref{tab:ablation-blip}, we demonstrate the result on BLIP for the image captioning task. As shown in Tab.~\ref{tab:ablation-blip}, merging only in the last block gives the best performance, which confirms our observation on ViT models. Note that due to the acceleration effect on the vision encoder, the speed-up effects vary slightly on BLIP. However, in MLLMs, the time consumption of the vision encoder is 10-15$\times$ inferior to LLM, thus yielding almost no difference for different merging positions.

\vspace{0.1cm}
\noindent \textbf{Aggregation Method.}
In Sec.~\ref{sec:aggregation}, we explore three distinct aggregation methods: Dropping, Average, and Weighted Average based on vector norms. As illustrated in Tab.~\ref{tab:ablation}, both average and weighted average merging show a clear advantage over direct dropping. This result is expected, as directly dropping tokens leads to greater information loss. To reduce module complexity, we used averaging in all MLLM experiments, as the differences are minimal.
\begin{table}[t]
\centering
\footnotesize
\caption{\textbf{Ablation Study on Aggregation Operation.} On BLIP~\cite{li2022blip} captioning task, we test the performance of 3 aggregation operations discussed in Sec.~\ref{sec:aggregation}. For fair comparison, we adopt the aggregation only in the last block.}
\vspace{-0.3cm}
\begin{tabular}{c|c|ccc}
\toprule
Method & \begin{tabular}[c]{@{}c@{}}Reduct \\ Ratio\end{tabular} & B@4 & CIDEr & SPICE \\
\hline \hline
Original & 0\% & 133.3 & 39.7 & 23.8 \\ \hline
Avg & \multirow{3}{*}{50\%} & 132.0 & 39.0 & \textbf{23.6} \\
Drop &  & 130.0 & 38.7 & 23.3 \\
Weighted &  & \textbf{132.2} & \textbf{39.1} & 23.6  \\
\hline
Avg & \multirow{3}{*}{60\%} & 131.0 & 38.9 & \textbf{23.5}  \\
Drop &  & 128.9 & 38.2 & 23.1 \\
Weighted &  & \textbf{131.5} & \textbf{39.1} & 23.4 \\

\bottomrule
\end{tabular}
\label{tab:ablation}
\vspace{-0.1cm}
\end{table}

\section{Conclusion}
In this paper, we introduce FOLDER, a plug-and-play module designed to efficiently reduce visual token sequences in MLLMs. Through empirical analysis of token reduction process, we develop a strategy named FOLDER that elaborately focuses on large-scale token reduction in last network layers while optimizing information retention. Our experiments demonstrate that FOLDER serves as a dual-purpose accelerator for MLLMs: it achieves up to 70\% reduction in visual tokens while delivering significant speedup factors of 1.8$\times$ for inference and 1.65$\times$ for training. Notably, FOLDER not only maintains but often enhances model performance across both inference and training scenarios, even for complicated tasks such as video understanding. The method's effectiveness across various architectures and tasks demonstrates its potential as a practical solution for making MLLMs more computationally efficient without compromising their capabilities.

{
\small
\bibliographystyle{ieeenat_fullname}
\bibliography{main}

\begin{thebibliography}{89}
\providecommand{\natexlab}[1]{#1}
\providecommand{\url}[1]{\texttt{#1}}
\expandafter\ifx\csname urlstyle\endcsname\relax
  \providecommand{\doi}[1]{doi: #1}\else
  \providecommand{\doi}{doi: \begingroup \urlstyle{rm}\Url}\fi

\bibitem[Achiam et~al.(2023)Achiam, Adler, Agarwal, Ahmad, Akkaya, Aleman,
  Almeida, Altenschmidt, Altman, Anadkat, et~al.]{achiam2023gpt}
Josh Achiam, Steven Adler, Sandhini Agarwal, Lama Ahmad, Ilge Akkaya,
  Florencia~Leoni Aleman, Diogo Almeida, Janko Altenschmidt, Sam Altman,
  Shyamal Anadkat, et~al.
\newblock Gpt-4 technical report.
\newblock \emph{arXiv preprint arXiv:2303.08774}, 2023.

\bibitem[Anthropic(2024)]{claude3}
Anthropic.
\newblock The claude 3 model family: Opus, sonnet, haiku, 2024.

\bibitem[Ataallah et~al.(2024)Ataallah, Shen, Abdelrahman, Sleiman, Zhuge,
  Ding, Zhu, Schmidhuber, and Elhoseiny]{ataallah2024MiniGPT4-Video}
Kirolos Ataallah, Xiaoqian Shen, Eslam Abdelrahman, Essam Sleiman, Mingchen
  Zhuge, Jian Ding, Deyao Zhu, Jürgen Schmidhuber, and Mohamed Elhoseiny.
\newblock Goldfish: Vision-language understanding of arbitrarily long videos,
  2024.

\bibitem[Block et~al.(2023)Block, Foster, Krishnamurthy, Simchowitz, and
  Zhang]{block2023butterfly}
Adam Block, Dylan~J Foster, Akshay Krishnamurthy, Max Simchowitz, and Cyril
  Zhang.
\newblock Butterfly effects of sgd noise: Error amplification in behavior
  cloning and autoregression.
\newblock \emph{arXiv preprint arXiv:2310.11428}, 2023.

\bibitem[Bolya et~al.(2022)Bolya, Fu, Dai, Zhang, Feichtenhofer, and
  Hoffman]{bolya2022token}
Daniel Bolya, Cheng-Yang Fu, Xiaoliang Dai, Peizhao Zhang, Christoph
  Feichtenhofer, and Judy Hoffman.
\newblock Token merging: Your vit but faster.
\newblock \emph{arXiv preprint arXiv:2210.09461}, 2022.

\bibitem[Chen et~al.(2023)Chen, Zhu, Shen, Li, Liu, Zhang, Krishnamoorthi,
  Chandra, Xiong, and Elhoseiny]{chen2023minigpt}
Jun Chen, Deyao Zhu, Xiaoqian Shen, Xiang Li, Zechun Liu, Pengchuan Zhang,
  Raghuraman Krishnamoorthi, Vikas Chandra, Yunyang Xiong, and Mohamed
  Elhoseiny.
\newblock Minigpt-v2: large language model as a unified interface for
  vision-language multi-task learning.
\newblock \emph{arXiv preprint arXiv:2310.09478}, 2023.

\bibitem[Chen et~al.(2024{\natexlab{a}})Chen, Li, Dong, Zhang, Zang, Chen,
  Duan, Wang, Qiao, Lin, et~al.]{chen2024we}
Lin Chen, Jinsong Li, Xiaoyi Dong, Pan Zhang, Yuhang Zang, Zehui Chen, Haodong
  Duan, Jiaqi Wang, Yu Qiao, Dahua Lin, et~al.
\newblock Are we on the right way for evaluating large vision-language models?
\newblock \emph{arXiv preprint arXiv:2403.20330}, 2024{\natexlab{a}}.

\bibitem[Chen et~al.(2024{\natexlab{b}})Chen, Zhao, Liu, Bai, Lin, Zhou, and
  Chang]{chen2024image}
Liang Chen, Haozhe Zhao, Tianyu Liu, Shuai Bai, Junyang Lin, Chang Zhou, and
  Baobao Chang.
\newblock An image is worth 1/2 tokens after layer 2: Plug-and-play inference
  acceleration for large vision-language models.
\newblock \emph{arXiv preprint arXiv:2403.06764}, 2024{\natexlab{b}}.

\bibitem[Chen et~al.(2024{\natexlab{c}})Chen, Chen, Zhai, Ju, Hong, Lan, and
  Xiao]{chen2024wear}
Mengting Chen, Xi Chen, Zhonghua Zhai, Chen Ju, Xuewen Hong, Jinsong Lan, and
  Shuai Xiao.
\newblock Wear-any-way: Manipulable virtual try-on via sparse correspondence
  alignment.
\newblock \emph{arXiv preprint arXiv:2403.12965}, 2024{\natexlab{c}}.

\bibitem[Chen et~al.(2024{\natexlab{d}})Chen, Cheng, Yao, Ju, Huang, Lan, Zeng,
  and Xiao]{chen2023enhancing}
Xu Chen, Zida Cheng, Jiangchao Yao, Chen Ju, Weilin Huang, Jinsong Lan, Xiaoyi
  Zeng, and Shuai Xiao.
\newblock Enhancing cross-domain click-through rate prediction via explicit
  feature augmentation.
\newblock In \emph{Int. World Wide Web Conf.}, 2024{\natexlab{d}}.

\bibitem[Chen et~al.(2024{\natexlab{e}})Chen, Wu, Wang, Su, Chen, Xing, Zhong,
  Zhang, Zhu, Lu, et~al.]{chen2024internvl}
Zhe Chen, Jiannan Wu, Wenhai Wang, Weijie Su, Guo Chen, Sen Xing, Muyan Zhong,
  Qinglong Zhang, Xizhou Zhu, Lewei Lu, et~al.
\newblock Internvl: Scaling up vision foundation models and aligning for
  generic visual-linguistic tasks.
\newblock In \emph{Proceedings of the IEEE/CVF Conference on Computer Vision
  and Pattern Recognition}, pages 24185--24198, 2024{\natexlab{e}}.

\bibitem[Cheng et~al.(2024)Cheng, Ju, Wang, Liu, Chen, Hu, Zhang, and
  Wang]{cheng2024denoiser}
Haozhe Cheng, Cheng Ju, Haicheng Wang, Jinxiang Liu, Mengting Chen, Qiang Hu,
  Xiaoyun Zhang, and Yanfeng Wang.
\newblock Denoiser: Rethinking the robustness for open-vocabulary action
  recognition.
\newblock \emph{arXiv preprint arXiv:2404.14890}, 2024.

\bibitem[Cheng et~al.(2023{\natexlab{a}})Cheng, Song, Ma, Zhu, Zhu, and
  Zhang]{cheng2023beyond}
Kanzhi Cheng, Wenpo Song, Zheng Ma, Wenhao Zhu, Zixuan Zhu, and Jianbing Zhang.
\newblock Beyond generic: Enhancing image captioning with real-world knowledge
  using vision-language pre-training model.
\newblock In \emph{Proceedings of the 31st ACM International Conference on
  Multimedia}, pages 5038--5047, 2023{\natexlab{a}}.

\bibitem[Cheng et~al.(2023{\natexlab{b}})Cheng, Ju, Chen, Zhai, Xiao, Zeng, and
  Huang]{cheng2023image}
Zida Cheng, Chen Ju, Xu Chen, Zhonghua Zhai, Shuai Xiao, Xiaoyi Zeng, and
  Weilin Huang.
\newblock Image to multi-modal retrieval for industrial scenarios.
\newblock \emph{arXiv preprint arXiv:2305.03972}, 2023{\natexlab{b}}.

\bibitem[Cheng et~al.(2023{\natexlab{c}})Cheng, Xiao, Zhai, Zeng, and
  Huang]{cheng2023mixer}
Zida Cheng, Shuai Xiao, Zhonghua Zhai, Xiaoyi Zeng, and Weilin Huang.
\newblock Mixer: Image to multi-modal retrieval learning for industrial
  application.
\newblock \emph{arXiv preprint arXiv:2305.03972}, 2023{\natexlab{c}}.

\bibitem[Dao(2023)]{dao2023flashattention2}
Tri Dao.
\newblock Flashattention-2: Faster attention with better parallelism and work
  partitioning, 2023.

\bibitem[Dao et~al.(2022)Dao, Fu, Ermon, Rudra, and Ré]{dao2022flashattention}
Tri Dao, Daniel~Y. Fu, Stefano Ermon, Atri Rudra, and Christopher Ré.
\newblock Flashattention: Fast and memory-efficient exact attention with
  io-awareness, 2022.

\bibitem[Deng et~al.(2009)Deng, Dong, Socher, Li, Li, and
  Fei-Fei]{deng2009imagenet}
Jia Deng, Wei Dong, Richard Socher, Li-Jia Li, Kai Li, and Li Fei-Fei.
\newblock Imagenet: A large-scale hierarchical image database.
\newblock In \emph{2009 IEEE conference on computer vision and pattern
  recognition}, pages 248--255. Ieee, 2009.

\bibitem[Dosovitskiy(2020)]{dosovitskiy2020image}
Alexey Dosovitskiy.
\newblock An image is worth 16x16 words: Transformers for image recognition at
  scale.
\newblock \emph{arXiv preprint arXiv:2010.11929}, 2020.

\bibitem[Eckart and Young(1936)]{eckart1936approximation}
Carl Eckart and Gale Young.
\newblock The approximation of one matrix by another of lower rank.
\newblock \emph{Psychometrika}, 1\penalty0 (3):\penalty0 211--218, 1936.

\bibitem[Fang et~al.(2024)Fang, Mao, Duan, Zhao, Li, Lin, and
  Chen]{fang2024mmbenchvideo}
Xinyu Fang, Kangrui Mao, Haodong Duan, Xiangyu Zhao, Yining Li, Dahua Lin, and
  Kai Chen.
\newblock Mmbench-video: A long-form multi-shot benchmark for holistic video
  understanding.
\newblock \emph{arXiv preprint arXiv:2406.14515}, 2024.

\bibitem[Fang et~al.(2021)Fang, Wang, Hu, Wang, Yang, and
  Liu]{fang2021compressing}
Zhiyuan Fang, Jianfeng Wang, Xiaowei Hu, Lijuan Wang, Yezhou Yang, and Zicheng
  Liu.
\newblock Compressing visual-linguistic model via knowledge distillation.
\newblock In \emph{ICCV}, 2021.

\bibitem[Frantar et~al.(2022)Frantar, Ashkboos, Hoefler, and
  Alistarh]{frantar2022gptq}
Elias Frantar, Saleh Ashkboos, Torsten Hoefler, and Dan Alistarh.
\newblock Gptq: Accurate post-training quantization for generative pre-trained
  transformers.
\newblock \emph{arXiv preprint arXiv:2210.17323}, 2022.

\bibitem[Fu et~al.(2023)Fu, Chen, Shen, Qin, Zhang, Lin, Yang, Zheng, Li, Sun,
  et~al.]{fu2023mme}
Chaoyou Fu, Peixian Chen, Yunhang Shen, Yulei Qin, Mengdan Zhang, Xu Lin,
  Jinrui Yang, Xiawu Zheng, Ke Li, Xing Sun, et~al.
\newblock Mme: A comprehensive evaluation benchmark for multimodal large
  language models.
\newblock \emph{arXiv preprint arXiv:2306.13394}, 2023.

\bibitem[Fu et~al.(2024)Fu, Dai, Luo, Li, Ren, Zhang, Wang, Zhou, Shen, Zhang,
  et~al.]{fu2024video}
Chaoyou Fu, Yuhan Dai, Yongdong Luo, Lei Li, Shuhuai Ren, Renrui Zhang, Zihan
  Wang, Chenyu Zhou, Yunhang Shen, Mengdan Zhang, et~al.
\newblock Video-mme: The first-ever comprehensive evaluation benchmark of
  multi-modal llms in video analysis.
\newblock \emph{arXiv preprint arXiv:2405.21075}, 2024.

\bibitem[Grok(2024)]{grok}
Grok.
\newblock Grok1.5 homepage.
\newblock \url{https://x.ai/blog/grok-1.5v}, 2024.
\newblock Accessed: 2024-11-14.

\bibitem[Guan et~al.(2024)Guan, Liu, Wu, Xian, Li, Liu, Wang, Chen, Huang,
  Yacoob, et~al.]{guan2024hallusionbench}
Tianrui Guan, Fuxiao Liu, Xiyang Wu, Ruiqi Xian, Zongxia Li, Xiaoyu Liu, Xijun
  Wang, Lichang Chen, Furong Huang, Yaser Yacoob, et~al.
\newblock Hallusionbench: an advanced diagnostic suite for entangled language
  hallucination and visual illusion in large vision-language models.
\newblock In \emph{Proceedings of the IEEE/CVF Conference on Computer Vision
  and Pattern Recognition}, pages 14375--14385, 2024.

\bibitem[He et~al.(2022)He, Chen, Xie, Li, Doll{\'a}r, and
  Girshick]{he2022masked}
Kaiming He, Xinlei Chen, Saining Xie, Yanghao Li, Piotr Doll{\'a}r, and Ross
  Girshick.
\newblock Masked autoencoders are scalable vision learners.
\newblock In \emph{Proceedings of the IEEE/CVF conference on computer vision
  and pattern recognition}, pages 16000--16009, 2022.

\bibitem[Huang et~al.(2024)Huang, Zhai, Shen, Cao, Zhao, Xu, Ye, and
  Lin]{huang2024dynamic}
Wenxuan Huang, Zijie Zhai, Yunhang Shen, Shaoshen Cao, Fei Zhao, Xiangfeng Xu,
  Zheyu Ye, and Shaohui Lin.
\newblock Dynamic-llava: Efficient multimodal large language models via dynamic
  vision-language context sparsification.
\newblock \emph{arXiv preprint arXiv:2412.00876}, 2024.

\bibitem[Jin et~al.(2024)Jin, Takanobu, Zhang, Cao, and Yuan]{jin2024chat}
Peng Jin, Ryuichi Takanobu, Wancai Zhang, Xiaochun Cao, and Li Yuan.
\newblock Chat-univi: Unified visual representation empowers large language
  models with image and video understanding.
\newblock In \emph{Proceedings of the IEEE/CVF Conference on Computer Vision
  and Pattern Recognition}, pages 13700--13710, 2024.

\bibitem[Ju et~al.(2020)Ju, Zhao, Zhang, Wang, and Tian]{ju2020point}
Chen Ju, Peisen Zhao, Ya Zhang, Yanfeng Wang, and Qi Tian.
\newblock Point-level temporal action localization: Bridging fully-supervised
  proposals to weakly-supervised losses.
\newblock \emph{arXiv preprint arXiv:2012.08236}, 2020.

\bibitem[Ju et~al.(2021)Ju, Zhao, Chen, Zhang, Wang, and Tian]{ju2021divide}
Chen Ju, Peisen Zhao, Siheng Chen, Ya Zhang, Yanfeng Wang, and Qi Tian.
\newblock Divide and conquer for single-frame temporal action localization.
\newblock In \emph{ICCV}, 2021.

\bibitem[Ju et~al.(2022{\natexlab{a}})Ju, Han, Zheng, Zhang, and
  Xie]{ju2022prompting}
Chen Ju, Tengda Han, Kunhao Zheng, Ya Zhang, and Weidi Xie.
\newblock Prompting visual-language models for efficient video understanding.
\newblock In \emph{ECCV}. Springer, 2022{\natexlab{a}}.

\bibitem[Ju et~al.(2022{\natexlab{b}})Ju, Zhao, Chen, Zhang, Zhang, and
  Tian]{ju2021adaptive}
Chen Ju, Peisen Zhao, Siheng Chen, Ya Zhang, Xiaoyun Zhang, and Qi Tian.
\newblock Adaptive mutual supervision for weakly-supervised temporal action
  localization.
\newblock \emph{IEEE TMM}, 2022{\natexlab{b}}.

\bibitem[Ju et~al.(2023{\natexlab{a}})Ju, Li, Zhao, Zhang, Zhang, Tian, Wang,
  and Xie]{ju2023multi}
Chen Ju, Zeqian Li, Peisen Zhao, Ya Zhang, Xiaopeng Zhang, Qi Tian, Yanfeng
  Wang, and Weidi Xie.
\newblock Multi-modal prompting for low-shot temporal action localization.
\newblock \emph{arXiv preprint arXiv:2303.11732}, 2023{\natexlab{a}}.

\bibitem[Ju et~al.(2023{\natexlab{b}})Ju, Wang, Li, Chen, Zhai, Huang, and
  Xiao]{ju2023turbo}
Chen Ju, Haicheng Wang, Zeqian Li, Xu Chen, Zhonghua Zhai, Weilin Huang, and
  Shuai Xiao.
\newblock Turbo: Informativity-driven acceleration plug-in for vision-language
  models.
\newblock \emph{arXiv preprint arXiv:2312.07408}, 2023{\natexlab{b}}.

\bibitem[Ju et~al.(2023{\natexlab{c}})Ju, Wang, Liu, Ma, Zhang, Zhao, Chang,
  and Tian]{ju2023constraint}
Chen Ju, Haicheng Wang, Jinxiang Liu, Chaofan Ma, Ya Zhang, Peisen Zhao,
  Jianlong Chang, and Qi Tian.
\newblock Constraint and union for partially-supervised temporal sentence
  grounding.
\newblock \emph{arXiv preprint arXiv:2302.09850}, 2023{\natexlab{c}}.

\bibitem[Ju et~al.(2023{\natexlab{d}})Ju, Zheng, Liu, Zhao, Zhang, Chang, Tian,
  and Wang]{ju2023distilling}
Chen Ju, Kunhao Zheng, Jinxiang Liu, Peisen Zhao, Ya Zhang, Jianlong Chang, Qi
  Tian, and Yanfeng Wang.
\newblock Distilling vision-language pre-training to collaborate with
  weakly-supervised temporal action localization.
\newblock In \emph{CVPR}, 2023{\natexlab{d}}.

\bibitem[Ju et~al.(2025)Ju, Wang, Cheng, Chen, Zhai, Huang, Lan, Xiao, and
  Zheng]{ju2025turbo}
Chen Ju, Haicheng Wang, Haozhe Cheng, Xu Chen, Zhonghua Zhai, Weilin Huang,
  Jinsong Lan, Shuai Xiao, and Bo Zheng.
\newblock Turbo: Informativity-driven acceleration plug-in for vision-language
  large models.
\newblock In \emph{European Conference on Computer Vision}, pages 436--455.
  Springer, 2025.

\bibitem[Kar et~al.(2024)Kar, Tonioni, Poklukar, Kulshrestha, Zamir, and
  Tombari]{kar2024brave}
O{\u{g}}uzhan~Fatih Kar, Alessio Tonioni, Petra Poklukar, Achin Kulshrestha,
  Amir Zamir, and Federico Tombari.
\newblock Brave: Broadening the visual encoding of vision-language models.
\newblock \emph{arXiv preprint arXiv:2404.07204}, 2024.

\bibitem[Karp et~al.(1990)Karp, Vazirani, and Vazirani]{karp1990optimal}
Richard~M Karp, Umesh~V Vazirani, and Vijay~V Vazirani.
\newblock An optimal algorithm for on-line bipartite matching.
\newblock In \emph{Proceedings of the twenty-second annual ACM symposium on
  Theory of computing}, pages 352--358, 1990.

\bibitem[Khan et~al.(2023)Khan, BG, Schulter, Yu, Fu, and
  Chandraker]{khan2023q}
Zaid Khan, Vijay~Kumar BG, Samuel Schulter, Xiang Yu, Yun Fu, and Manmohan
  Chandraker.
\newblock Q: How to specialize large vision-language models to data-scarce vqa
  tasks? a: Self-train on unlabeled images!
\newblock In \emph{Proceedings of the IEEE/CVF Conference on Computer Vision
  and Pattern Recognition}, pages 15005--15015, 2023.

\bibitem[Kong et~al.(2022)Kong, Dong, Ma, Meng, Sun, Niu, Shen, Yuan, Ren, Qin,
  Tang, and Wang]{kong2022spvitenablingfastervision}
Zhenglun Kong, Peiyan Dong, Xiaolong Ma, Xin Meng, Mengshu Sun, Wei Niu, Xuan
  Shen, Geng Yuan, Bin Ren, Minghai Qin, Hao Tang, and Yanzhi Wang.
\newblock Spvit: Enabling faster vision transformers via soft token pruning,
  2022.

\bibitem[Li et~al.(2023{\natexlab{a}})Li, Wang, Wang, Ge, Ge, and
  Shan]{li2023seed}
Bohao Li, Rui Wang, Guangzhi Wang, Yuying Ge, Yixiao Ge, and Ying Shan.
\newblock Seed-bench: Benchmarking multimodal llms with generative
  comprehension.
\newblock \emph{arXiv preprint arXiv:2307.16125}, 2023{\natexlab{a}}.

\bibitem[Li et~al.(2022)Li, Li, Xiong, and Hoi]{li2022blip}
Junnan Li, Dongxu Li, Caiming Xiong, and Steven Hoi.
\newblock Blip: Bootstrapping language-image pre-training for unified
  vision-language understanding and generation.
\newblock In \emph{International conference on machine learning}, pages
  12888--12900. PMLR, 2022.

\bibitem[Li et~al.(2023{\natexlab{b}})Li, Li, Savarese, and Hoi]{li2023blip}
Junnan Li, Dongxu Li, Silvio Savarese, and Steven Hoi.
\newblock Blip-2: Bootstrapping language-image pre-training with frozen image
  encoders and large language models.
\newblock In \emph{International conference on machine learning}, pages
  19730--19742. PMLR, 2023{\natexlab{b}}.

\bibitem[Li et~al.(2023{\natexlab{c}})Li, He, Wang, Li, Wang, Luo, Wang, Wang,
  and Qiao]{li2023videochat}
KunChang Li, Yinan He, Yi Wang, Yizhuo Li, Wenhai Wang, Ping Luo, Yali Wang,
  Limin Wang, and Yu Qiao.
\newblock Videochat: Chat-centric video understanding.
\newblock \emph{arXiv preprint arXiv:2305.06355}, 2023{\natexlab{c}}.

\bibitem[Li et~al.(2019)Li, Yatskar, Yin, Hsieh, and Chang]{visualbert}
Liunian~Harold Li, Mark Yatskar, Da Yin, Cho{-}Jui Hsieh, and Kai{-}Wei Chang.
\newblock Visualbert: {A} simple and performant baseline for vision and
  language, 2019.

\bibitem[Liang et~al.(2022)Liang, Ge, Tong, Song, Wang, and
  Xie]{liang2022patchesneedexpeditingvision}
Youwei Liang, Chongjian Ge, Zhan Tong, Yibing Song, Jue Wang, and Pengtao Xie.
\newblock Not all patches are what you need: Expediting vision transformers via
  token reorganizations, 2022.

\bibitem[Lin et~al.(2023)Lin, Ye, Zhu, Cui, Ning, Jin, and Yuan]{lin2023video}
Bin Lin, Yang Ye, Bin Zhu, Jiaxi Cui, Munan Ning, Peng Jin, and Li Yuan.
\newblock Video-llava: Learning united visual representation by alignment
  before projection.
\newblock \emph{arXiv preprint arXiv:2311.10122}, 2023.

\bibitem[Lin et~al.(2014)Lin, Maire, Belongie, Hays, Perona, Ramanan,
  Doll{\'a}r, and Zitnick]{lin2014microsoft}
Tsung-Yi Lin, Michael Maire, Serge Belongie, James Hays, Pietro Perona, Deva
  Ramanan, Piotr Doll{\'a}r, and C~Lawrence Zitnick.
\newblock Microsoft coco: Common objects in context.
\newblock In \emph{Computer Vision--ECCV 2014: 13th European Conference,
  Zurich, Switzerland, September 6-12, 2014, Proceedings, Part V 13}, pages
  740--755. Springer, 2014.

\bibitem[Liu et~al.(2023{\natexlab{a}})Liu, Li, Wu, and Lee]{llava}
Haotian Liu, Chunyuan Li, Qingyang Wu, and Yong~Jae Lee.
\newblock Visual instruction tuning.
\newblock In \emph{NeurIPS}, 2023{\natexlab{a}}.

\bibitem[Liu et~al.(2023{\natexlab{b}})Liu, Zaharia, and
  Abbeel]{liu2023ringAttention}
Hao Liu, Matei Zaharia, and Pieter Abbeel.
\newblock Ring attention with blockwise transformers for near-infinite context,
  2023{\natexlab{b}}.

\bibitem[Liu et~al.(2024{\natexlab{a}})Liu, Li, Li, and Lee]{liu2024improved}
Haotian Liu, Chunyuan Li, Yuheng Li, and Yong~Jae Lee.
\newblock Improved baselines with visual instruction tuning.
\newblock In \emph{Proceedings of the IEEE/CVF Conference on Computer Vision
  and Pattern Recognition}, pages 26296--26306, 2024{\natexlab{a}}.

\bibitem[Liu et~al.(2022)Liu, Ju, Xie, and Zhang]{liu2022exploiting}
Jinxiang Liu, Chen Ju, Weidi Xie, and Ya Zhang.
\newblock Exploiting transformation invariance and equivariance for
  self-supervised sound localisation.
\newblock In \emph{ACM MM}, 2022.

\bibitem[Liu et~al.(2023{\natexlab{c}})Liu, Ju, Ma, Wang, Wang, and
  Zhang]{liu2023audio}
Jinxiang Liu, Chen Ju, Chaofan Ma, Yanfeng Wang, Yu Wang, and Ya Zhang.
\newblock Audio-aware query-enhanced transformer for audio-visual segmentation.
\newblock \emph{arXiv preprint arXiv:2307.13236}, 2023{\natexlab{c}}.

\bibitem[Liu et~al.(2024{\natexlab{b}})Liu, Liu, Zhang, Ju, Zhang, and
  Wang]{liu2024audio}
Jinxiang Liu, Yikun Liu, Fei Zhang, Chen Ju, Ya Zhang, and Yanfeng Wang.
\newblock Audio-visual segmentation via unlabeled frame exploitation.
\newblock \emph{arXiv preprint arXiv:2403.11074}, 2024{\natexlab{b}}.

\bibitem[Liu et~al.(2024{\natexlab{c}})Liu, Wang, Ju, Ma, Zhang, and
  Xie]{liu2024annotation}
Jinxiang Liu, Yu Wang, Chen Ju, Chaofan Ma, Ya Zhang, and Weidi Xie.
\newblock Annotation-free audio-visual segmentation.
\newblock In \emph{Proceedings of the IEEE/CVF Winter Conference on
  Applications of Computer Vision}, 2024{\natexlab{c}}.

\bibitem[Lu et~al.(2022)Lu, Mishra, Xia, Qiu, Chang, Zhu, Tafjord, Clark, and
  Kalyan]{lu2022learn}
Pan Lu, Swaroop Mishra, Tony Xia, Liang Qiu, Kai-Wei Chang, Song-Chun Zhu,
  Oyvind Tafjord, Peter Clark, and Ashwin Kalyan.
\newblock Learn to explain: Multimodal reasoning via thought chains for science
  question answering.
\newblock In \emph{The 36th Conference on Neural Information Processing Systems
  (NeurIPS)}, 2022.

\bibitem[Ma et~al.(2023{\natexlab{a}})Ma, Yang, Ju, Zhang, Liu, Wang, Zhang,
  and Wang]{ma2023diffusionseg}
Chaofan Ma, Yuhuan Yang, Chen Ju, Fei Zhang, Jinxiang Liu, Yu Wang, Ya Zhang,
  and Yanfeng Wang.
\newblock Diffusionseg: Adapting diffusion towards unsupervised object
  discovery.
\newblock \emph{arXiv preprint arXiv:2303.09813}, 2023{\natexlab{a}}.

\bibitem[Ma et~al.(2023{\natexlab{b}})Ma, Yang, Ju, Zhang, Zhang, and
  Wang]{ma2023attrseg}
Chaofan Ma, Yuhuan Yang, Chen Ju, Fei Zhang, Ya Zhang, and Yanfeng Wang.
\newblock Attrseg: open-vocabulary semantic segmentation via attribute
  decomposition-aggregation.
\newblock \emph{arXiv preprint arXiv:2309.00096}, 2023{\natexlab{b}}.

\bibitem[Ma et~al.(2023{\natexlab{c}})Ma, Yang, Ju, Zhang, Zhang, and
  Wang]{ma2023open}
Chaofan Ma, Yuhuan Yang, Chen Ju, Fei Zhang, Ya Zhang, and Yanfeng Wang.
\newblock Open-vocabulary semantic segmentation via attribute
  decomposition-aggregation.
\newblock \emph{arXiv preprint arXiv:2309.00096}, 2023{\natexlab{c}}.

\bibitem[Ma et~al.(2024)Ma, Yang, Ju, Zhang, Zhang, and Wang]{ma2024open}
Chaofan Ma, Yuhuan Yang, Chen Ju, Fei Zhang, Ya Zhang, and Yanfeng Wang.
\newblock Open-vocabulary semantic segmentation via attribute
  decomposition-aggregation.
\newblock \emph{NeurIPS}, 2024.

\bibitem[Mishra et~al.(2019)Mishra, Shekhar, Singh, and
  Chakraborty]{mishraICDAR19}
Anand Mishra, Shashank Shekhar, Ajeet~Kumar Singh, and Anirban Chakraborty.
\newblock Ocr-vqa: Visual question answering by reading text in images.
\newblock In \emph{ICDAR}, 2019.

\bibitem[Oquab et~al.(2023)Oquab, Darcet, Moutakanni, Vo, Szafraniec, Khalidov,
  Fernandez, Haziza, Massa, El-Nouby, et~al.]{oquab2023dinov2}
Maxime Oquab, Timoth{\'e}e Darcet, Th{\'e}o Moutakanni, Huy Vo, Marc
  Szafraniec, Vasil Khalidov, Pierre Fernandez, Daniel Haziza, Francisco Massa,
  Alaaeldin El-Nouby, et~al.
\newblock Dinov2: Learning robust visual features without supervision.
\newblock \emph{arXiv preprint arXiv:2304.07193}, 2023.

\bibitem[Radford et~al.(2021)Radford, Kim, Hallacy, Ramesh, Goh, Agarwal,
  Sastry, Askell, Mishkin, Clark, et~al.]{radford2021learning}
Alec Radford, Jong~Wook Kim, Chris Hallacy, Aditya Ramesh, Gabriel Goh,
  Sandhini Agarwal, Girish Sastry, Amanda Askell, Pamela Mishkin, Jack Clark,
  et~al.
\newblock Learning transferable visual models from natural language
  supervision.
\newblock In \emph{International conference on machine learning}, pages
  8748--8763. PMLR, 2021.

\bibitem[Rao et~al.(2021)Rao, Zhao, Liu, Lu, Zhou, and
  Hsieh]{rao2021dynamicvit}
Yongming Rao, Wenliang Zhao, Benlin Liu, Jiwen Lu, Jie Zhou, and Cho-Jui Hsieh.
\newblock Dynamicvit: Efficient vision transformers with dynamic token
  sparsification.
\newblock \emph{NeurIPS}, 2021.

\bibitem[Rubner et~al.(2000)Rubner, Tomasi, and Guibas]{rubner2000earth}
Yossi Rubner, Carlo Tomasi, and Leonidas~J Guibas.
\newblock The earth mover's distance as a metric for image retrieval.
\newblock \emph{International journal of computer vision}, 40:\penalty0
  99--121, 2000.

\bibitem[Schwenk et~al.(2022)Schwenk, Khandelwal, Clark, Marino, and
  Mottaghi]{schwenk2022okvqa}
Dustin Schwenk, Apoorv Khandelwal, Christopher Clark, Kenneth Marino, and
  Roozbeh Mottaghi.
\newblock A-okvqa: A benchmark for visual question answering using world
  knowledge.
\newblock In \emph{European conference on computer vision}, pages 146--162.
  Springer, 2022.

\bibitem[Shang et~al.(2024)Shang, Cai, Xu, Lee, and Yan]{shang2024llava}
Yuzhang Shang, Mu Cai, Bingxin Xu, Yong~Jae Lee, and Yan Yan.
\newblock Llava-prumerge: Adaptive token reduction for efficient large
  multimodal models.
\newblock \emph{arXiv preprint arXiv:2403.15388}, 2024.

\bibitem[Shi et~al.(2023)Shi, Tao, Jin, Yang, Yuan, and Wang]{shi2023upop}
Dachuan Shi, Chaofan Tao, Ying Jin, Zhendong Yang, Chun Yuan, and Jiaqi Wang.
\newblock Upop: Unified and progressive pruning for compressing vision-language
  transformers.
\newblock \emph{arXiv preprint arXiv:2301.13741}, 2023.

\bibitem[Shi et~al.(2024)Shi, Liu, Wang, Liao, Radhakrishnan, Huang, Yin,
  Sapra, Yacoob, Shi, et~al.]{shi2024eagle}
Min Shi, Fuxiao Liu, Shihao Wang, Shijia Liao, Subhashree Radhakrishnan, De-An
  Huang, Hongxu Yin, Karan Sapra, Yaser Yacoob, Humphrey Shi, et~al.
\newblock Eagle: Exploring the design space for multimodal llms with mixture of
  encoders.
\newblock \emph{arXiv preprint arXiv:2408.15998}, 2024.

\bibitem[Singh et~al.(2022)Singh, Hu, Goswami, Couairon, Galuba, Rohrbach, and
  Kiela]{flava}
Amanpreet Singh, Ronghang Hu, Vedanuj Goswami, Guillaume Couairon, Wojciech
  Galuba, Marcus Rohrbach, and Douwe Kiela.
\newblock {FLAVA:} {A} foundational language and vision alignment model.
\newblock In \emph{Proceedings of the IEEE/CVF Conference on Computer Vision
  and Pattern Recognition}, 2022.

\bibitem[Tan and Bansal(2019)]{lxmert}
Hao Tan and Mohit Bansal.
\newblock {LXMERT}: Learning cross-modality encoder representations from
  transformers.
\newblock In \emph{Proceedings of the 2019 Conference on Empirical Methods in
  Natural Language Processing and the 9th International Joint Conference on
  Natural Language Processing, {EMNLP-IJCNLP}, November 3-7, 2019}, 2019.

\bibitem[Team et~al.(2023)Team, Anil, Borgeaud, Wu, Alayrac, Yu, Soricut,
  Schalkwyk, Dai, Hauth, et~al.]{team2023gemini}
Gemini Team, Rohan Anil, Sebastian Borgeaud, Yonghui Wu, Jean-Baptiste Alayrac,
  Jiahui Yu, Radu Soricut, Johan Schalkwyk, Andrew~M Dai, Anja Hauth, et~al.
\newblock Gemini: a family of highly capable multimodal models.
\newblock \emph{arXiv preprint arXiv:2312.11805}, 2023.

\bibitem[Tong et~al.(2024)Tong, Liu, Zhai, Ma, LeCun, and Xie]{tong2024eyes}
Shengbang Tong, Zhuang Liu, Yuexiang Zhai, Yi Ma, Yann LeCun, and Saining Xie.
\newblock Eyes wide shut? exploring the visual shortcomings of multimodal llms.
\newblock In \emph{Proceedings of the IEEE/CVF Conference on Computer Vision
  and Pattern Recognition}, pages 9568--9578, 2024.

\bibitem[Wang et~al.(2024{\natexlab{a}})Wang, Ju, Lin, Xiao, Chen, Huang, Liu,
  Yao, Lan, Chen, et~al.]{wang2024advancing}
Haicheng Wang, Chen Ju, Weixiong Lin, Shuai Xiao, Mengting Chen, Yixuan Huang,
  Chang Liu, Mingshuai Yao, Jinsong Lan, Ying Chen, et~al.
\newblock Advancing myopia to holism: Fully contrastive language-image
  pre-training.
\newblock \emph{arXiv preprint arXiv:2412.00440}, 2024{\natexlab{a}}.

\bibitem[Wang et~al.(2024{\natexlab{b}})Wang, Bai, Tan, Wang, Fan, Bai, Chen,
  Liu, Wang, Ge, et~al.]{wang2024qwen2}
Peng Wang, Shuai Bai, Sinan Tan, Shijie Wang, Zhihao Fan, Jinze Bai, Keqin
  Chen, Xuejing Liu, Jialin Wang, Wenbin Ge, et~al.
\newblock Qwen2-vl: Enhancing vision-language model's perception of the world
  at any resolution.
\newblock \emph{arXiv preprint arXiv:2409.12191}, 2024{\natexlab{b}}.

\bibitem[Wang et~al.(2022)Wang, Zhou, Zeng, and Zhang]{wang2022efficientvlm}
Tiannan Wang, Wangchunshu Zhou, Yan Zeng, and Xinsong Zhang.
\newblock Efficientvlm: Fast and accurate vision-language models via knowledge
  distillation and modal-adaptive pruning.
\newblock \emph{arXiv preprint arXiv:2210.07795}, 2022.

\bibitem[Xiao et~al.(2023)Xiao, Lin, Seznec, Wu, Demouth, and
  Han]{xiao2023smoothquant}
Guangxuan Xiao, Ji Lin, Mickael Seznec, Hao Wu, Julien Demouth, and Song Han.
\newblock Smoothquant: Accurate and efficient post-training quantization for
  large language models.
\newblock In \emph{International Conference on Machine Learning}, pages
  38087--38099. PMLR, 2023.

\bibitem[Yang et~al.(2024{\natexlab{a}})Yang, Chen, Tian, Wang, Li, Yu, and
  Jia]{yang2024visionzip}
Senqiao Yang, Yukang Chen, Zhuotao Tian, Chengyao Wang, Jingyao Li, Bei Yu, and
  Jiaya Jia.
\newblock Visionzip: Longer is better but not necessary in vision language
  models.
\newblock \emph{arXiv preprint arXiv:2412.04467}, 2024{\natexlab{a}}.

\bibitem[Yang et~al.(2023)Yang, Ma, Ju, Zhang, and Wang]{yang2023multi}
Yuhuan Yang, Chaofan Ma, Chen Ju, Ya Zhang, and Yanfeng Wang.
\newblock Multi-modal prototypes for open-set semantic segmentation.
\newblock \emph{arXiv preprint arXiv:2307.02003}, 2023.

\bibitem[Yang et~al.(2024{\natexlab{b}})Yang, Ma, Ju, Zhang, Yao, Zhang, and
  Wang]{yang2024multi}
Yuhuan Yang, Chaofan Ma, Chen Ju, Fei Zhang, Jiangchao Yao, Ya Zhang, and
  Yanfeng Wang.
\newblock Multi-modal prototypes for open-world semantic segmentation.
\newblock \emph{IJCV}, 2024{\natexlab{b}}.

\bibitem[Yifan et~al.(2023)Yifan, Yifan, Kun, Jinpeng, Wayne, and
  Ji-Rong]{Li-hallucination-2023}
Li Yifan, Du Yifan, Zhou Kun, Wang Jinpeng, Xin~Zhao Wayne, and Wen Ji-Rong.
\newblock Evaluating object hallucination in large vision-language models.
\newblock In \emph{The 2023 Conference on Empirical Methods in Natural Language
  Processing}, 2023.

\bibitem[Yuan et~al.(2023)Yuan, Haodong, Yuanhan, Bo, Songyang, Wangbo, Yike,
  Jiaqi, Conghui, Ziwei, Kai, and Dahua]{MMBench}
Liu Yuan, Duan Haodong, Zhang Yuanhan, Li Bo, Zhang Songyang, Zhao Wangbo, Yuan
  Yike, Wang Jiaqi, He Conghui, Liu Ziwei, Chen Kai, and Lin Dahua.
\newblock Mmbench: Is your multi-modal model an all-around player?
\newblock \emph{arXiv:2307.06281}, 2023.

\bibitem[Yue et~al.(2024)Yue, Ni, Zhang, Zheng, Liu, Zhang, Stevens, Jiang,
  Ren, Sun, Wei, Yu, Yuan, Sun, Yin, Zheng, Yang, Liu, Huang, Sun, Su, and
  Chen]{yue2023mmmu}
Xiang Yue, Yuansheng Ni, Kai Zhang, Tianyu Zheng, Ruoqi Liu, Ge Zhang, Samuel
  Stevens, Dongfu Jiang, Weiming Ren, Yuxuan Sun, Cong Wei, Botao Yu, Ruibin
  Yuan, Renliang Sun, Ming Yin, Boyuan Zheng, Zhenzhu Yang, Yibo Liu, Wenhao
  Huang, Huan Sun, Yu Su, and Wenhu Chen.
\newblock Mmmu: A massive multi-discipline multimodal understanding and
  reasoning benchmark for expert agi.
\newblock In \emph{Proceedings of CVPR}, 2024.

\bibitem[Zhang et~al.(2023)Zhang, Li, and Bing]{zhang2023Video-LLaMA}
Hang Zhang, Xin Li, and Lidong Bing.
\newblock Video-llama: An instruction-tuned audio-visual language model for
  video understanding, 2023.

\bibitem[Zhang et~al.(2024)Zhang, Fan, Ma, Zheng, Huang, Cheng, Gudovskiy,
  Okuno, Nakata, Keutzer, et~al.]{zhang2024sparsevlm}
Yuan Zhang, Chun-Kai Fan, Junpeng Ma, Wenzhao Zheng, Tao Huang, Kuan Cheng,
  Denis Gudovskiy, Tomoyuki Okuno, Yohei Nakata, Kurt Keutzer, et~al.
\newblock Sparsevlm: Visual token sparsification for efficient vision-language
  model inference.
\newblock \emph{arXiv preprint arXiv:2410.04417}, 2024.

\bibitem[Zhao et~al.(2020)Zhao, Xie, Ju, Zhang, Wang, and Tian]{zhao2020bottom}
Peisen Zhao, Lingxi Xie, Chen Ju, Ya Zhang, Yanfeng Wang, and Qi Tian.
\newblock Bottom-up temporal action localization with mutual regularization.
\newblock In \emph{ECCV}. Springer, 2020.

\end{thebibliography}
}

\clearpage
\appendix

\setcounter{page}{1}
In the supplementary material, we first provide more details about our FOLDER module and experiment setup in Sec.~\ref{sec:detail}. Then we offer more results on empirical studies (Sec.~\ref{subsec:emp}) as well as ablation studies including the propagation effect on MLLMs (Sec.~\ref{subsec: prop}) and the choice of aggregation strategies (Sec.~\ref{subsec: agg}). Finally we discuss about the limitations and future work (Sec.~\ref{sec:limit}).

\section{Implementation Details} \label{sec:detail}
\subsection{FOLDER Architecture for MLLMs}
FOLDER is designed as a plug-and-play module plugged  in transformer-based vision encoders of Multi-modal Large Language Models (MLLMs), to reduce the output visual token sequence length. FOLDER is a generalized token merging module, that can adapt any matching function or aggregation methods, allowing any number of tokens to be merged in any block without constraint. More specifically, we applied FOLDER between the residual connection of attention block and the MLP. To minimize the calculation overhead for token grouping, we upgrade the bipartite soft matching~\cite{karp1990optimal} algorithm to meet the demand for arbitrary number of merging.

\noindent \textbf{Merging Position} ToMe~\cite{bolya2022token} or Turbo~\cite{ju2025turbo} applies an uniform and progressive token merging, with constant number of reduction on each block to accelerate the vision encoder (ViT~\cite{dosovitskiy2020image}, CLIP~\cite{radford2021learning}, BLIP~\cite{li2022blip} etc.). Unlike them, we here focus on the acceleration of MLLMs, where the computational cost is concentrated in the LLM. Indeed, for LLaVA1.5-13B~\cite{liu2024improved}, the total time to generate the 1st-token is around 0.37s on V100-32G, while the vision encoder part is around 0.03s, which is less than $1/12$. For all the experiments in the main paper, we only reduce tokens in the output layer/block of vision encoder (LLaVA1.5~\cite{liu2024improved} uses the image feature of the second last layer, while Minigpt4v2~\cite{chen2023minigpt} uses only the last layer). The ablation on reduction partition between blocks conducted on MLLMs is provided in Tab.~\ref{tab:propagation}. Similar to the result on BLIP in the main paper, reducing tokens only in the last layer yields the best result, which is in line with our empirical observation.

\noindent \textbf{Matching Function} In FOLDER algorithm, we need to choose a matching function that can evaluate the similarity between tokens, so that we aggregate tokens with similar semantic meanings. ToMe~\cite{bolya2022token} directly calculates the cosine similarity between tokens' key value in attention calculation ($K$ taking the mean on multi-head), while Turbo~\cite{ju2025turbo} leverage a more delicate matching function that considers both similarity between tokens and the semantic importance of tokens (attention contribution for the class token). For the experiment in the main paper, we adopt the matching function of Turbo, by replacing the metric of key value by token itself. To minimize the implementation effort, we offer an extremely simplified version that only evaluate the cosine similarity between tokens in the last layer (between attention and MLP), and the performance gap is minor (please refer to the ablation studies in section~\ref{subsec: agg}). This allows the adaptation to be a ready-to-use on any MLLM visual backbones.

\noindent \textbf{Merging Order} To realize the average merging that is independent of the folding order, we make one little adjustment. For example, if we have two folding operations, which asks for token $(x_1,x_3,x_5)$ to be merged as $x_8$ in the first fold, and $(x_6,x_7,x_8)$ to be merged in the second fold. We would like to average on $(x_1,x_3,x_5,x_6,x_7)$, without taking the merging order into account. 
$$x_{\text{merge}}=\text{avg}(x_1,x_3,x_5,x_6,x_7)$$
To do this, we use a size list to note the number of tokens that contributed to obtaining the merged token (for the merged token $x_8$ after the first fold, the corresponding size is 3), then we weight the token by their size during the following fold (for token $x_8$, we considered $3\times x_8$ during the average computation with $(x_6,x_7)$: 
$$x_8=\text{avg}(x_1,x_3,x_5)$$
$$x_{\text{merge}}=(3 \times x_8+x_6+x_7)/(3+1+1)$$
In this way, we realize average merging regardless of the folding order. We release the code on BLIP and LLaVA. For more implementation details, please refer to the code along with this file.

\subsection{Training Details of LLaVA1.5-13B}
To test the effectiveness of FOLDER for training phase, we train LLaVA1.5-13B following the same procedures (pretrain-sft two-stage training, dataset, training parameters) detailed in LLaVA1.5 repository, with FOLDER inserted in the visual backbone of LLaVA1.5. The whole training process is conducted on 8 A100-80G GPUs and the GPU hour in Tab.~6 in the main paper is evaluated by the actual training time multiplying the number of GPUs.
\begin{figure}[t]
  \centering
  \includegraphics[width=\columnwidth]{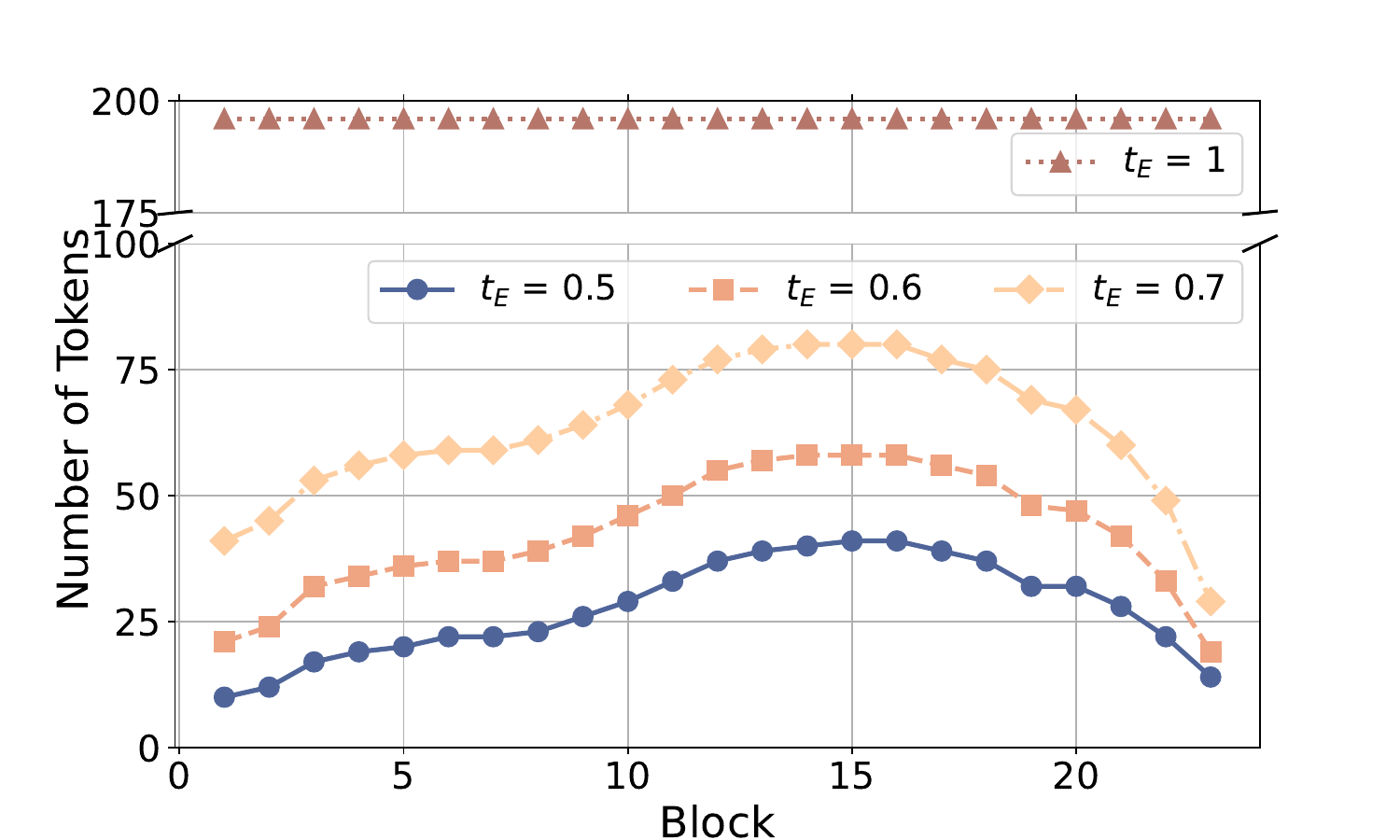}
  \caption{\textbf{Minimum Number of Tokens with Energy $t_{E}$ Across Blocks on ViT Large.} We evaluate on 3 different $t_{E}$ for every block on ViT large 24 blocks.}
  \label{fig:energy-large}
\end{figure}

\begin{figure}[t]
  \centering
  \includegraphics[width=\columnwidth]{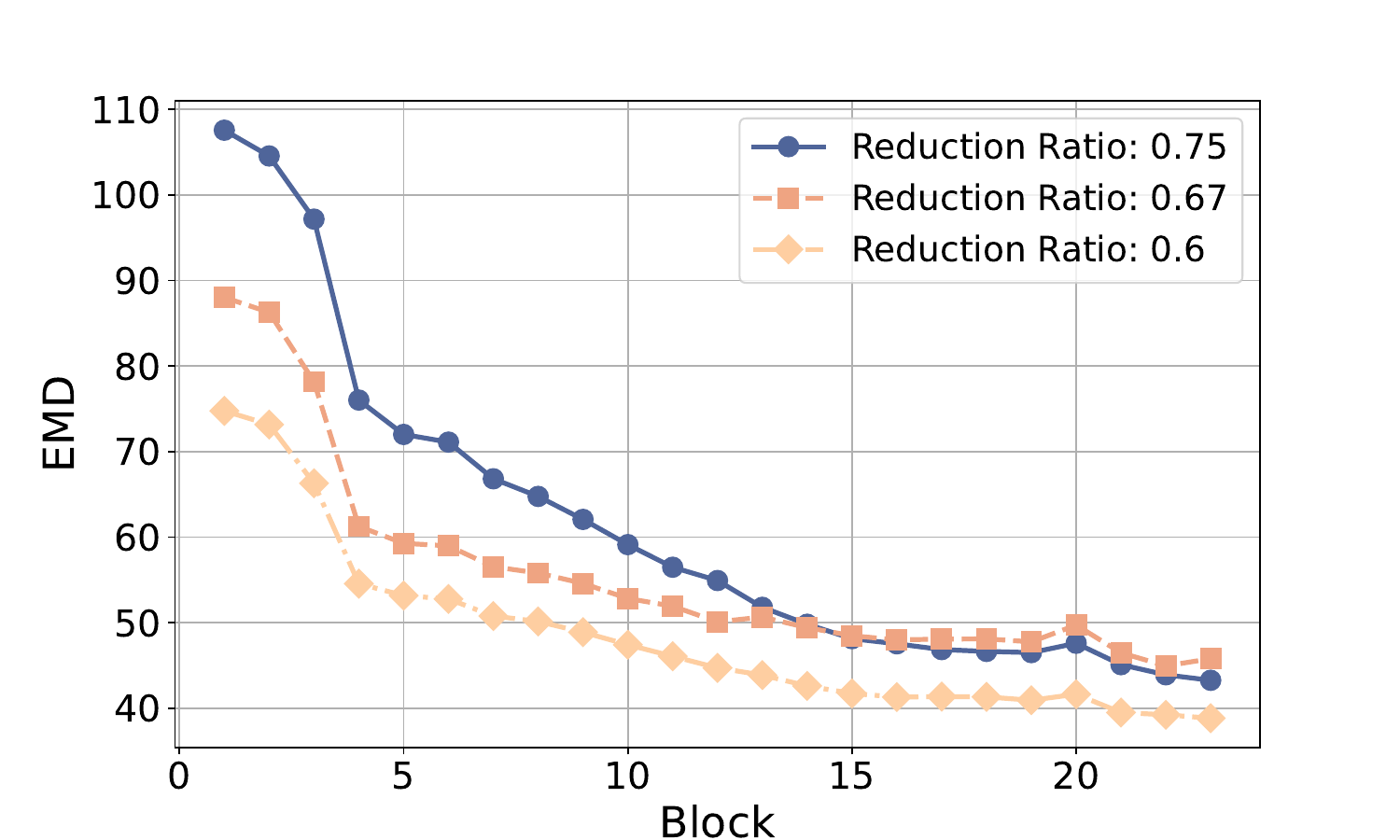}
  \caption{\textbf{EMD Distance Between Reduced and Original Output Distributions under 3 Reduction Ratios on ViT Large.} We compare the EMD distance by exerting token reduction on different blocks on ViT large 24 blocks.}
  \label{fig:emd-large}
\end{figure}

\begin{figure}[t]
  \centering
  \includegraphics[width=\columnwidth]{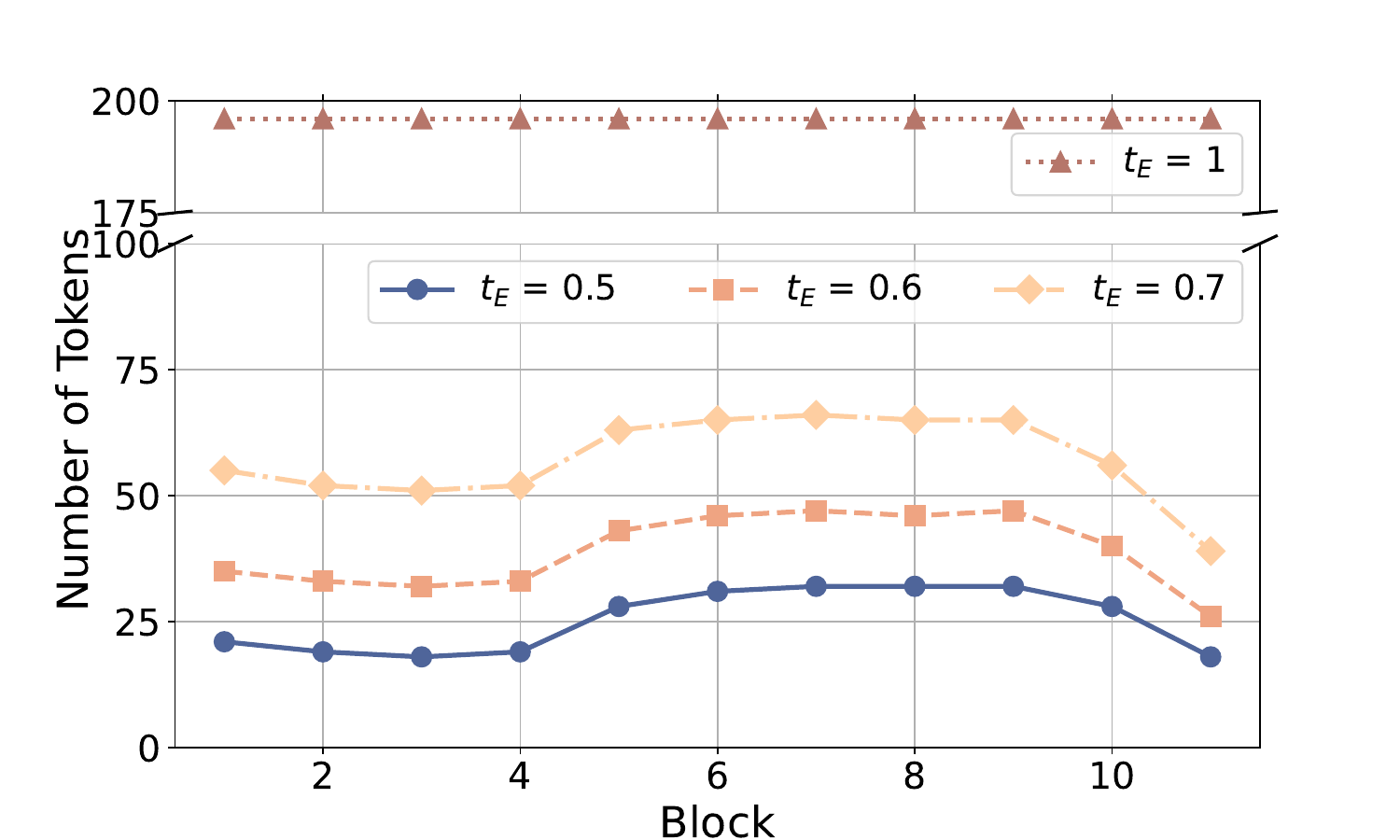}
  \caption{\textbf{Minimum Number of Tokens with Energy $t_{E}$ Across Blocks on ViT Small.} We evaluate on 3 different $t_{E}$ for every block on ViT small 12 blocks.}
  \label{fig:energy-small}
\end{figure}

\begin{figure}[t]
  \centering
  \includegraphics[width=\columnwidth]{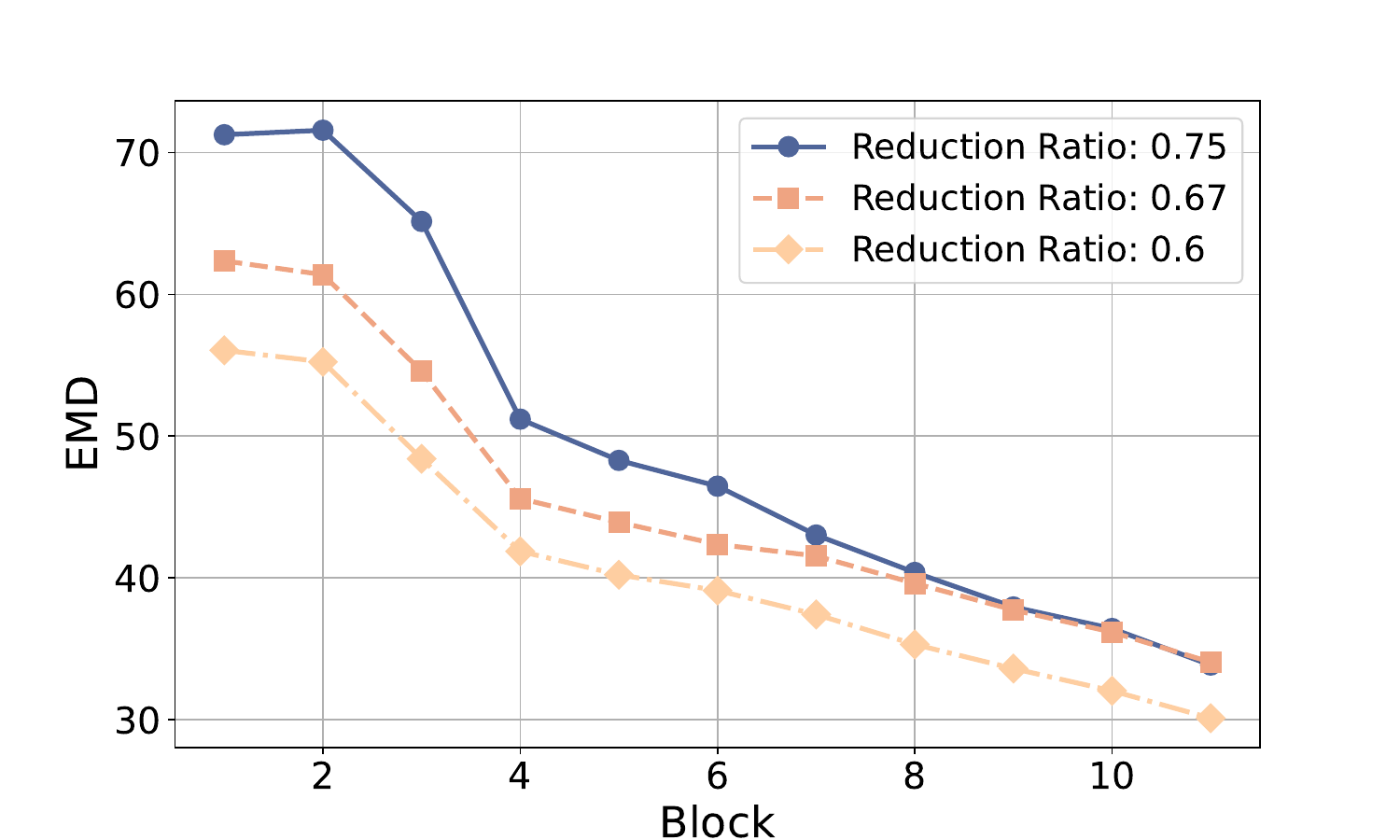}
  \caption{\textbf{EMD Distance Between Reduced and Original Output Distributions under 3 Reduction Ratios on ViT Small.} We compare the EMD distance by exerting token reduction on different blocks on ViT small 12 blocks.}
  \label{fig:emd-small}
\end{figure}

\begin{table*}[t]
\centering
\footnotesize
\caption{\textbf{Propagation Effect on LLaVA1.5 7B/13B.} We evaluate different merging positions using LLaVA1.5 under 66\% reduction ratio. Last-n refers to uniform reduction in last n blocks, uniform refers to uniform reduction in every block.}
\begin{tabular}{c|c|C{0.8cm}C{0.6cm}cC{0.8cm}ccccC{0.8cm}cc}
\hline
Method & \begin{tabular}[c]{@{}c@{}}Reduct \\ Ratio\end{tabular} & \begin{tabular}[c]{@{}c@{}}MMBench\\ EN\end{tabular} & \begin{tabular}[c]{@{}c@{}}MM\\ Star\end{tabular} & MME & \begin{tabular}[c]{@{}c@{}}A-OK\\ VQA\end{tabular} & \begin{tabular}[c]{@{}c@{}}Hallusion\\ Bench\end{tabular} & \begin{tabular}[c]{@{}c@{}}MM\\ MU\end{tabular} & \begin{tabular}[c]{@{}c@{}}OCR\\ VQA\end{tabular} & \begin{tabular}[c]{@{}c@{}}\scriptsize SEED\\ \scriptsize BenchIMG\end{tabular} & \begin{tabular}[c]{@{}c@{}}Science\\ QA\end{tabular} & POPE & Avg \\ \hline  \hline

Original-7B & 0\%& 62.8 & 32.7 & 1338.9 & 78.8 & 35.6 & 32.2 & 52.4 & 60.2 & 68.1 & 79.7 & 55.0 \\ \hline

Uniform & \multirow{4}{*}{66\%} & 60.1 & \textbf{31.5} & 1301.6 & \textbf{78.0} & 34.9 & 24.7 & 34.4 & 57.8 & 67.2 & 85.0 & 52.0 \\

Last-3 & & 60.9 & 31.1 & 1312.4 & 77.4 & 36.4 & \textbf{31.6} & 41.9 & 58.5 & 67.9 & 85.7 & 53.9\\

Last-2 & & 61.1 & 30.4 & \textbf{1353.1} & 76.9 & 35.9 & 31.4 & 41.6 & 58.4 & 67.9 & \textbf{85.7} & 53.8 \\

Last-1 & & \textbf{61.4} & 30.3 & 1350.0 & 77.9 & \textbf{39.2} & 31.3 & \textbf{46.1} & \textbf{59.7} & \textbf{68.3} & 85.4 & \textbf{54.8} \\
\hline \hline
Original-13B & 0\% & 66.6 & 30.9 & 1371.1 & 77.0 & 36.1 & 34.0 & 54.9 & 59.4 & 68.8 & 86.4 & 56.3\\ \hline
Uniform & \multirow{4}{*}{66\%} & 64.7 & 31.2 & 1168.3 & 76.3 & 25.9 & 33.3 & 47.9 & 58.5 & 70.5 & 85.5 & 53.6\\

Last-3 & & \textbf{66.5} & 31.6 & \textbf{1369.4} & 77.2 & 34.5 & 34.8 & 50.1 & 58.7 & \textbf{70.9} & \textbf{86.5} & 55.9\\

Last-2 & & 65.7 & \textbf{32.5} & 1332.0 & 77.2 & \textbf{34.6} & 34.5 & 51.7 & 58.6 & 70.3 & 86.0 & 55.9 \\

Last-1 &  & 65.8 & 31.7 & 1366.9 & \textbf{77.3} & 33.8 & \textbf{35.0} & \textbf{52.6} & \textbf{58.8} & 70.7 & 86.1 & \textbf{56.1} \\ 
\hline
\end{tabular}
\label{tab:propagation}
\end{table*}
\begin{table*}[t]
\centering
\footnotesize
\caption{\textbf{Aggregation method on LLaVA1.5 7B/13B.} We test the performance of 3 aggregation operations. For fair comparison, we adopt the aggregation only in the last block.}
\begin{tabular}{c|c|C{0.8cm}C{0.6cm}cC{0.8cm}ccccC{0.8cm}cc}
\hline
Method & \begin{tabular}[c]{@{}c@{}}Reduct \\ Ratio\end{tabular} & \begin{tabular}[c]{@{}c@{}}MMBench\\ EN\end{tabular} & \begin{tabular}[c]{@{}c@{}}MM\\ Star\end{tabular} & MME & \begin{tabular}[c]{@{}c@{}}A-OK\\ VQA\end{tabular} & \begin{tabular}[c]{@{}c@{}}Hallusion\\ Bench\end{tabular} & \begin{tabular}[c]{@{}c@{}}MM\\ MU\end{tabular} & \begin{tabular}[c]{@{}c@{}}OCR\\ VQA\end{tabular} & \begin{tabular}[c]{@{}c@{}}\scriptsize SEED\\ \scriptsize BenchIMG\end{tabular} & \begin{tabular}[c]{@{}c@{}}Science\\ QA\end{tabular} & POPE & Avg \\ \hline  \hline
Original-7B & 0\%& 62.8 & 32.7 & 1338.9 & 78.8 & 35.6 & 32.2 & 52.4 & 60.2 & 68.1 & 79.7 & 55.0 \\ \hline
Direct Drop & \multirow{3}{*}{66\%} & 39.1 & 20.7 & 980.5 & 64.4 & 20.1 & 14.0 & 11.5 & 45.6 & 45.3 & 77.9 & 37.4 \\
Weighted Avg & & 60.5 & \textbf{30.9} & 1332.3 & 77.7 & 38.8 & 31.3 & \textbf{46.6} & \textbf{59.8} & 68.2 & \textbf{85.6} & 54.7\\
Avg &  & \textbf{61.4} & 30.3 & \textbf{1350.0} & \textbf{77.9} & \textbf{39.2} & \textbf{31.3} & \textbf{46.1} & 59.7 & \textbf{68.3} & 85.4 & \textbf{54.8} \\
\hline \hline
Original-13B & 0\% & 66.6 & 30.9 & 1371.1 & 77.0 & 36.1 & 34.0 & 54.9 & 59.4 & 68.8 & 86.4 & 56.3\\ \hline
Direct Drop & \multirow{3}{*}{66\%} & 43.1 & 20.6 & 1082.6 & 65.5 & 18.2 & 19.3 & 29.9 & 45.8 & 47.3 & 79.5 & 40.8\\
Weighted Avg & & 65.3 & 31.3 & \textbf{1374.2} & \textbf{77.4} & \textbf{35.0} & 34.6 & 52.5 & 58.3 & 70.3 & 85.8 & 56.0\\
Avg &  & \textbf{65.8} & \textbf{31.7} & 1366.9 & 77.3 & 33.8 & \textbf{35.0} & \textbf{52.6} & \textbf{58.8} & \textbf{70.7} & \textbf{86.1} & \textbf{56.1} \\ 
\hline
\end{tabular}
\label{tab:aggregation}
\end{table*}

\begin{table*}[t]
\centering
\footnotesize
\caption{\textbf{Ablation for Matching Functions on LLaVA1.5 7B/13B.} We evaluate various matching functions evolved from ToMe~\cite{bolya2022token} and Turbo~\cite{ju2025turbo}. The default setting is $\alpha=5$ in Eq.~\ref{eq:turbo} and metric as token itself. $\text{Metric} = K$ means that we use the key value to calculate cosine similarity between tokens.}
\begin{tabular}{c|c|C{0.8cm}C{0.6cm}cC{0.8cm}ccccC{0.8cm}cc}
\hline
Method & \begin{tabular}[c]{@{}c@{}}Reduct \\ Ratio\end{tabular} & \begin{tabular}[c]{@{}c@{}}MMBench\\ EN\end{tabular} & \begin{tabular}[c]{@{}c@{}}MM\\ Star\end{tabular} & MME & \begin{tabular}[c]{@{}c@{}}A-OK\\ VQA\end{tabular} & \begin{tabular}[c]{@{}c@{}}Hallusion\\ Bench\end{tabular} & \begin{tabular}[c]{@{}c@{}}MM\\ MU\end{tabular} & \begin{tabular}[c]{@{}c@{}}OCR\\ VQA\end{tabular} & \begin{tabular}[c]{@{}c@{}}\scriptsize SEED\\ \scriptsize BenchIMG\end{tabular} & \begin{tabular}[c]{@{}c@{}}Science\\ QA\end{tabular} & POPE & Avg \\ \hline  \hline

Original-7B & 0\%& 62.8 & 32.7 & 1338.9 & 78.8 & 35.6 & 32.2 & 52.4 & 60.2 & 68.1 & 79.7 & 55.0 \\ \hline

$\alpha=0$ & \multirow{4}{*}{66\%} & 60.9 & 30.7 & 1354.1 & 76.9 & 39.2 & 31.5 & 45.3 & 59.8 & 68.0 & \textbf{85.9} & 54.7 \\

$\alpha=3$ & & 60.8 & \textbf{30.7} & 1341.4 & 77.7 & 39.1 & 31.2 & \textbf{46.4} & \textbf{59.8} & \textbf{68.5} & 85.7 & 54.8 \\

$\alpha=5$ & & \textbf{61.4} & 30.3 & 1350.0 & 77.9 & 39.2 & 31.3 & 46.1 & 59.7 & 68.3 & 85.4 & 54.8 \\

$\text{Metric}=K$ & & 61.3 & 30.3 & \textbf{1354.9} & \textbf{78.1} & \textbf{39.4} & \textbf{32.4} & 45.2 & 59.6 & 68.3 & 85.3 & \textbf{54.8} \\
\hline \hline
Original-13B & 0\% & 66.6 & 30.9 & 1371.1 & 77.0 & 36.1 & 34.0 & 54.9 & 59.4 & 68.8 & 86.4 & 56.3\\ \hline
$\alpha=0$ & \multirow{4}{*}{66\%} & 65.3 & 31.7 & 1351.4 & \textbf{77.9} & 33.9 & \textbf{35.2} & 52.5 & 58.5 & 70.7 & 86.1 & 56.0 \\

$\alpha=3$ & & 65.1 & \textbf{31.9} & \textbf{1383.9} & 77.4 & \textbf{34.7} & 34.8 & \textbf{52.6} & 58.4 & 70.4 & 86.0 & 56.1 \\

$\alpha=5$ & & \textbf{65.8} & 31.7 & 1366.9 & 77.3 & 33.8 & 35.0 & 52.6 & \textbf{58.8} & 70.7 & \textbf{86.1} & \textbf{56.1} \\

$\text{Metric}=K$ &  & 65.4 & 31.7 & 1359.5 & 77.8 & 33.1 & 34.0 & 51.4 & 58.2 & \textbf{70.9} & 85.4 & 55.6 \\ 
\hline
\end{tabular}
\label{tab:alpha}
\end{table*}
\begin{table*}[t]
\centering
\footnotesize
\caption{\textbf{Simplified Version of FOLDER on LLaVA1.5 7B/13B.} By directly applying FOLDER on the output of visual encoder, we achieve similar performance with respect to standard FOLDER, while greatly reduce the implementation effort.}
\begin{tabular}{c|c|C{0.8cm}C{0.6cm}cC{0.8cm}ccccC{0.8cm}cc}
\hline
Method & \begin{tabular}[c]{@{}c@{}}Reduct \\ Ratio\end{tabular} & \begin{tabular}[c]{@{}c@{}}MMBench\\ EN\end{tabular} & \begin{tabular}[c]{@{}c@{}}MM\\ Star\end{tabular} & MME & \begin{tabular}[c]{@{}c@{}}A-OK\\ VQA\end{tabular} & \begin{tabular}[c]{@{}c@{}}Hallusion\\ Bench\end{tabular} & \begin{tabular}[c]{@{}c@{}}MM\\ MU\end{tabular} & \begin{tabular}[c]{@{}c@{}}OCR\\ VQA\end{tabular} & \begin{tabular}[c]{@{}c@{}}\scriptsize SEED\\ \scriptsize BenchIMG\end{tabular} & \begin{tabular}[c]{@{}c@{}}Science\\ QA\end{tabular} & POPE & Avg \\ \hline  \hline

Original-7B & 0\%& 62.8 & 32.7 & 1338.9 & 78.8 & 35.6 & 32.2 & 52.4 & 60.2 & 68.1 & 79.7 & 55.0 \\ \hline

Simplified & \multirow{2}{*}{66\%} & 60.7 & \textbf{30.8} & 1341.1 & 76.4 & 38.8 & \textbf{31.6} & \textbf{46.4} & \textbf{60.1} & \textbf{68.4} & \textbf{86.3} & 54.7 \\

Ours-standard &  & \textbf{61.4} & 30.3 & \textbf{1350.0} & \textbf{77.9} & \textbf{39.2} & 31.3 & 46.1 & 59.7 & 68.3 & 85.4 & \textbf{54.8} \\
\hline \hline
Original-13B & 0\% & 66.6 & 30.9 & 1371.1 & 77.0 & 36.1 & 34.0 & 54.9 & 59.4 & 68.8 & 86.4 & 56.3\\ \hline

Simplified & \multirow{2}{*}{66\%} & 65.4 & \textbf{31.9} & 1357.9 & \textbf{77.7} & \textbf{34.3} & 34.7 & 52.2 & 58.6 & \textbf{70.8} & 86.1 & 56.0\\

Ours-standard &  & \textbf{65.8} & 31.7 & \textbf{1366.9} & 77.3 & 33.8 & \textbf{35.0} & \textbf{52.6} & \textbf{58.8} & 70.7 & \textbf{86.1} & \textbf{56.1} \\ 
\hline
\end{tabular}
\label{tab:simple}
\end{table*}
\section{More Experiments}
\subsection{Results on Empirical Studies} \label{subsec:emp}
In addition to the empirical results on ViT-B, we also conduct such experiments on ViT-S and ViT-L to demonstrate the generality of such phenomenon. As shown in Fig.~\ref{fig:energy-large}, ~\ref{fig:emd-large},~\ref{fig:energy-small} and~\ref{fig:emd-small}, on models of various sizes, the trend of Energy \& EMD distance with respect to blocks is similar. Combined with the results in Tab.~\ref{tab:propagation}, we can conclude that merging on last layers is the best choice.

\subsection{Ablation Study on Propagation Effect} \label{subsec: prop}
In Tab.~8 of the main paper, we study the propagation effect on BLIP~\cite{li2022blip}. In Tab.~\ref{tab:propagation}, we offer results on LLaVA1.5-13B, with 4 different reduction partitions (keep the number of remaining tokens unchanged). Reduction in the last layer remains the best strategy for MLLMs of different sizes. Although more subtle partitions can be explored, merging only in the last layer is a simple and safe choice.

\subsection{Ablation Study on Aggregation Strategy} \label{subsec: agg}
\noindent \textbf{Aggregation Method. } In Tab.~\ref{tab:aggregation}, we evaluate the three aggregation methods on MLLMs discussed in the main paper. More specifically, we conduct the experiment on LLaVA1.5-7B and 13B to fully compare these aggregation methods. As shown in Tab.~\ref{tab:aggregation}, there is a significant reduction in performance using direct drop, especially on dense tasks like OCRVQA. This suggests that direct drop may cause severe information loss. While for merging strategies, weighted average on norm and vanilla average merging both shows superior performance over direct dropping. The performance gap between is minor. For simplicity, we use average merging in as our default aggregation method.

\noindent\textbf{Matching Function. } In addition to the aggregation method, we also conduct experiments on matching functions. Turbo~\cite{ju2025turbo} proposed a generalized matching function that considers both mutual redundancy (token similarity) and semantic values (attention importance), which is formulated as:
\begin{equation}
    \mathcal{E} = \mathcal{R}-\alpha\mathcal{I},
    \label{eq:turbo}
\end{equation}
where $\mathcal{R}$ the similarity between tokens and $\mathcal{I}$ token's attention contribution with respect to the class token. $\alpha$ is a weighted hyper-parameter which we take $\alpha=5$ (a rough approximation for $\alpha=\text{seq\_len}//100$). To calculate the similarity between tokens, ToMe and Turbo~\cite{ju2025turbo} leverage the $K$ (key) in the attention by taking mean on multi-head dimension, thus to save computational cost and enhance slightly the performance. In our experiment, we simply take the token itself as the metric to calculate the cosine similarity for simplicity. In Tab.~\ref{tab:alpha}, we ablate on different matching functions. By using various $\alpha$ values and metrics on LLaVA1.5-7B and 13B, the performance rests similar, indicating the robustness of FOLDER.

Furthermore, in order to avoid all potential difficulties in implementing FOLDER ({\em e.g.}, FOLDER needs to be inserted into vision backbone, which requires to adapt for different vision encoder's architecture), we introduce one extremely simplified version that only applies FOLDER on the output visual sequence from vision backbone, regardless of the model architecture. In Tab.~\ref{tab:simple}, we show that such a simplified version achieves almost the same performance as standard ones, further showing the stability and universality of our proposed method.

\section{Limitations \& Future Work} \label{sec:limit}
Due to the limited resources on Openai API, for all the MLLM benchmarks (except VQA tasks that is mandatory to use LLM evaluation), we adopt exact matching, which can result in a slight degradation of actual performance evaluated by LLM. However, it's still fair for the comparison. What's more, the mechanism in the performance boost remains unclear, as well as a more theoretical interpretation on token reduction is yet to explore.

\end{document}